\documentclass{macro/neu_msthesis}

\usepackage[style=numeric, doi=false, url=false, isbn=false,sorting=none]{biblatex}
\usepackage{url}
\usepackage{booktabs}
\usepackage{tabularx}
\usepackage{graphics}
\usepackage{biblatex}
\usepackage{multicol}
\usepackage{xcolor}
\title{ Integration of Polyimide Flexible PCB Wings in Northeastern's Aerobat}

\author{Yizhe Xu}

\nuid{001580560}

\dept{Mechanical and Industrial Engineering}

\degreename{Mechanical Engineering}

\field{Mechanical Engineering}

\submitdate{August 2023}

\numberofmembers{3}
\principaladviser{Prof. Alireza Ramezani}
\firstreader{Prof. Sipahi, Rifat }

\chairman{Prof. Marilyn L. Minus}
\dean{Dr. Waleed Meleis}

\usepackage{amsfonts,amssymb,amsmath}			
\usepackage{times}		
\usepackage{bm}
\usepackage{subfig}

\newcommand{\ifno}[1]{}

\usepackage{multirow}
\makeindex
\usepackage{makeidx}
\usepackage{acronym}

\usepackage{url}
\ifnum\pdfoutput>0
\usepackage[pdftex]{graphicx}
\usepackage{epstopdf}
\else
\usepackage{graphicx}
\fi

\ifnum\pdfoutput>0
\usepackage[pdftex,
bookmarks=true,
bookmarksnumbered=true,
hypertexnames=false,
breaklinks=true           
]{hyperref}
\else
\usepackage[hypertex,
bookmarks=true,
bookmarksnumbered=true,
hypertexnames=false,
breaklinks=true           
]{hyperref}
\fi

\hypersetup{				
pdfauthor = {\authorRef},
pdftitle = {\titleRef},
pdfsubject = {\expandafter{\degreeRef} thesis submitted to Northeastern University},
pdfkeywords = {add keywords here}
pdfcreator = {LaTeX with hyperref package},
pdfproducer = {dvips + ps2pdf}}

\usepackage{microtype}

\begin{document}

\pdfbookmark[1]{Cover}{cover}

\titlepage

\begin{frontmatter}


\begin{dedication}
To my family.
\end{dedication}


\pdfbookmark[1]{Table of Contents}{contents}
\tableofcontents
\listoffigures
\newpage\ssp
\listoftables


\chapter*{List of Acronyms}
\addcontentsline{toc}{chapter}{List of Acronyms}

\begin{acronym}
\item DoF - Degree of Freedom.
\item MAV - Micro Aerial Vehicle.
\item UAS - Unmanned Aerial System.
\item KS - Kinetic Sculpture.
\item IMU - Inertial Measurement Unit.
\item PWM - Pulse Width Modulation.
\item ESC - Electrical Speed Controller.
\item LiPo - Lithium-polymer.


\end{acronym}


\begin{acknowledgements}

I would like to take this opportunity to express my deepest gratitude to everyone who has supported and encouraged me throughout my thesis journey.

First and foremost, I would like to extend my heartfelt thanks to my advisor, Professor Alireza Ramezani, whose invaluable guidance, trust, and patience have allowed me to focus on overcoming challenges and fostering personal growth. His consistent presence in the lab and active engagement in experiments have been a source of immense inspiration and motivation. I am also grateful to my co-advisor, Professor Sipahi Rifat, for his insightful feedback on my thesis and for serving on my thesis committee.

Additionally, I would like to express my appreciation to my fellow SiliconSynapse lab mates, including, Bibek Gupta, Aniket Dhole, Xuejian Niu, Adarsh Salagame, Chenghao Wang, Shreyansh Pitroda, and Kaushik Venkatesh Krishnamurthy, for their collaborative spirit and unwavering encouragement. Their willingness to participate in experiments and brainstorm together has fostered an environment of camaraderie and intellectual curiosity.

In conclusion, I feel truly fortunate to have been surrounded by such a dedicated and supportive network of mentors, colleagues, and friends during this important phase of my academic journey. Their collective contributions have made this thesis not only possible but also a truly enriching experience.

\end{acknowledgements}


\begin{abstract}

The principal aim of this Master's thesis is to propel the optimization of the membrane wing structure of the Northeastern Aerobat through origami techniques and enhancing its capacity for secure hovering within confined spaces. Bio-inspired drones offer distinctive capabilities that pave the way for innovative applications, encompassing wildlife monitoring, precision agriculture, search and rescue operations, as well as the augmentation of residential safety. The evolved noise-reduction mechanisms of birds and insects prove advantageous for drones utilized in tasks like surveillance and wildlife observation, ensuring operation devoid of disturbances. Traditional flying drones equipped with rotary or fixed wings encounter notable constraints when navigating narrow pathways. While rotary and fixed-wing systems are conventionally harnessed for surveillance and reconnaissance, the integration of onboard sensor suites within micro aerial vehicles (MAVs) has garnered interest in vigilantly monitoring hazardous scenarios in residential settings. Notwithstanding the agility and commendable fault tolerance exhibited by systems such as quadrotors in demanding conditions, their inflexible body structures impede collision tolerance, necessitating operational spaces free of collisions. Recent years have witnessed an upsurge in integrating soft and pliable materials into the design of such systems; however, the pursuit of aerodynamic efficiency curtails the utilization of excessively flexible materials for rotor blades or propellers. This thesis introduces a design that integrates polyimide flexible PCBs into the wings of the Aerobat and employs guard design incorporating feedback-driven stabilizers, enabling stable hovering flights within Northeastern's Robotics-Inspired Study and Experimentation (RISE) cage.

\end{abstract}

\end{frontmatter}

\pagestyle{headings}


\chapter{Motivation and Literature Review}
\label{chap:intro}

The primary objective of this work is to take major steps toward helping Northeastern Aerobat \cite{sihite2022unsteady} 
hover safely inside confined spaces. Bioinspired drones such as Aerobat hold significant potential. It can significantly benefit various aspects of our lives and society, including environmental monitoring, search and rescue operations for trapped individuals, and enhancing residential safety\cite{pavlidis_urban_2001,everaerts2008use}.

\begin{figure}[htp]
    \centering
    \includegraphics[width=12cm, height=18cm]{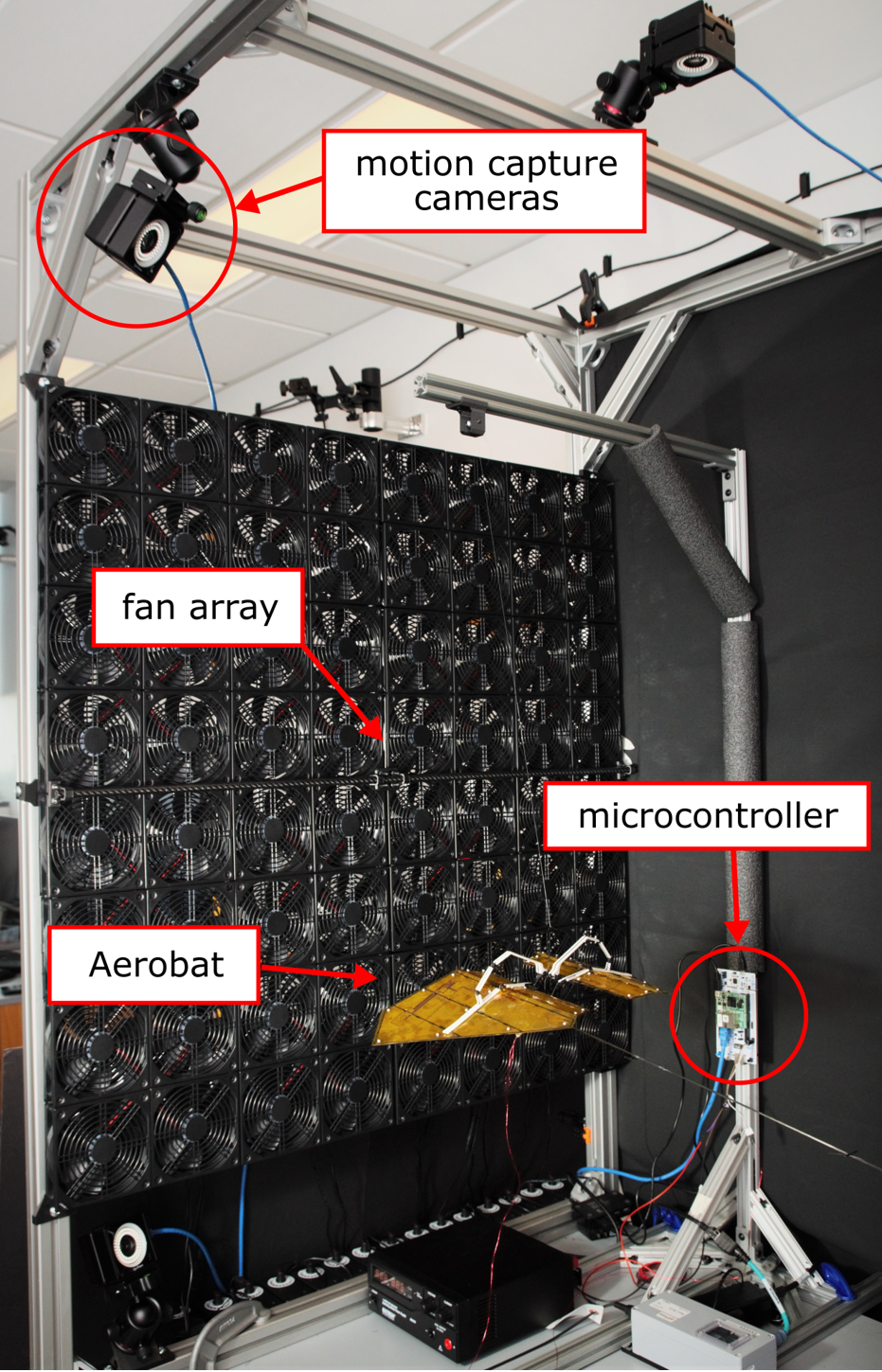}
    \caption{Illustrates Robotics-Inspired Study and Experimentation (RISE) cage at Northeastern University designed to explore flight in confined spaces. The MS thesis aims to integrate flexible polyimide in Aerobat's design and help Aerobat \cite{sihite2022unsteady} safely hover in this space. }
    \label{fig:rise}
\end{figure}

\begin{figure}[htp]
    \centering
    \includegraphics[width=1\linewidth]{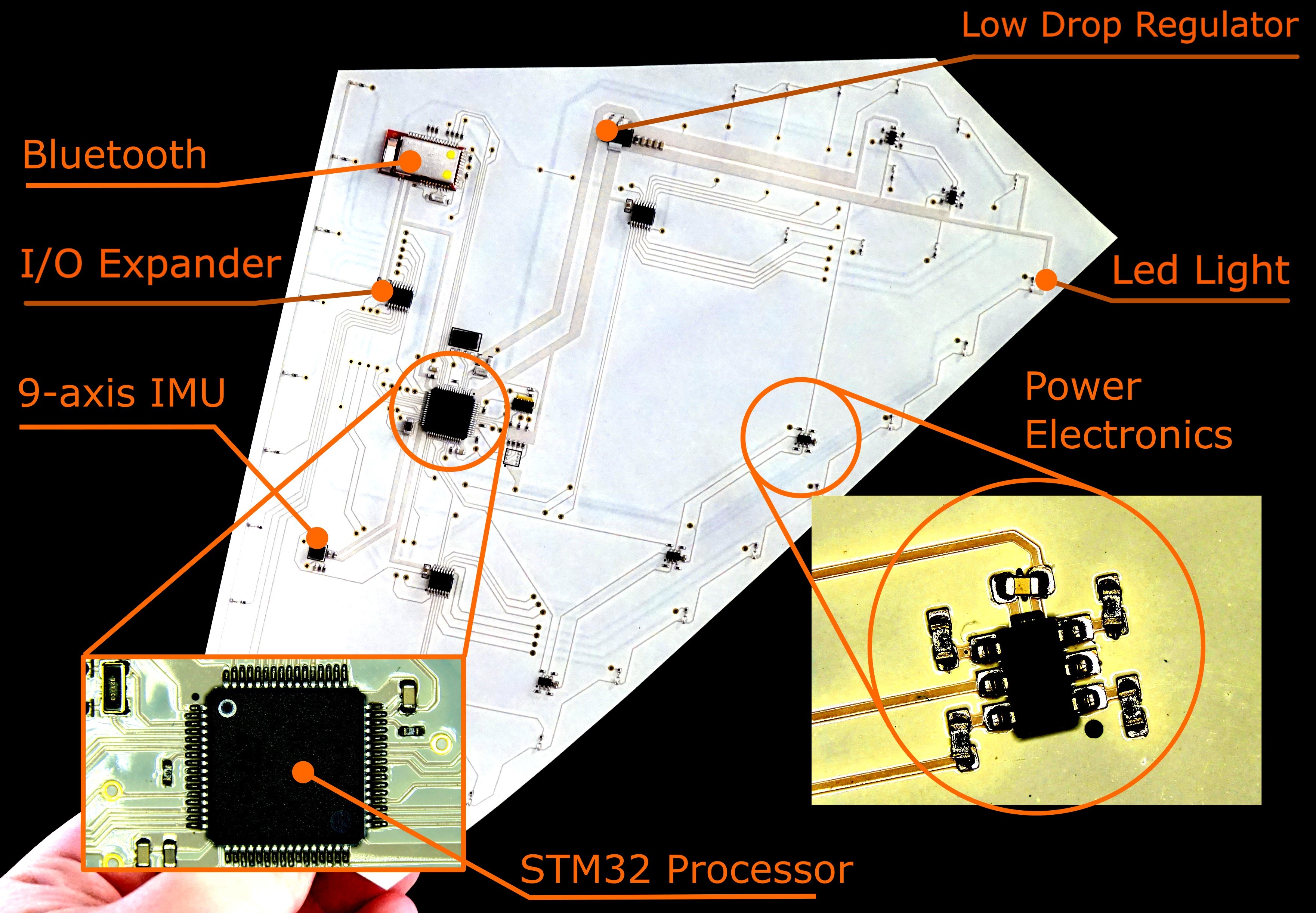}
    \caption{Detailed view of polyimide flexible PCB }
    \label{fig:FlexPcb}
\end{figure}

\renewcommand\floatpagefraction{.9}
\renewcommand\topfraction{.9}
\renewcommand\bottomfraction{.9}
\renewcommand\textfraction{.1}
\setcounter{totalnumber}{50}
\setcounter{topnumber}{50}
\setcounter{bottomnumber}{50}

\begin{figure}
    \centering
    \includegraphics[width=1.0\linewidth]{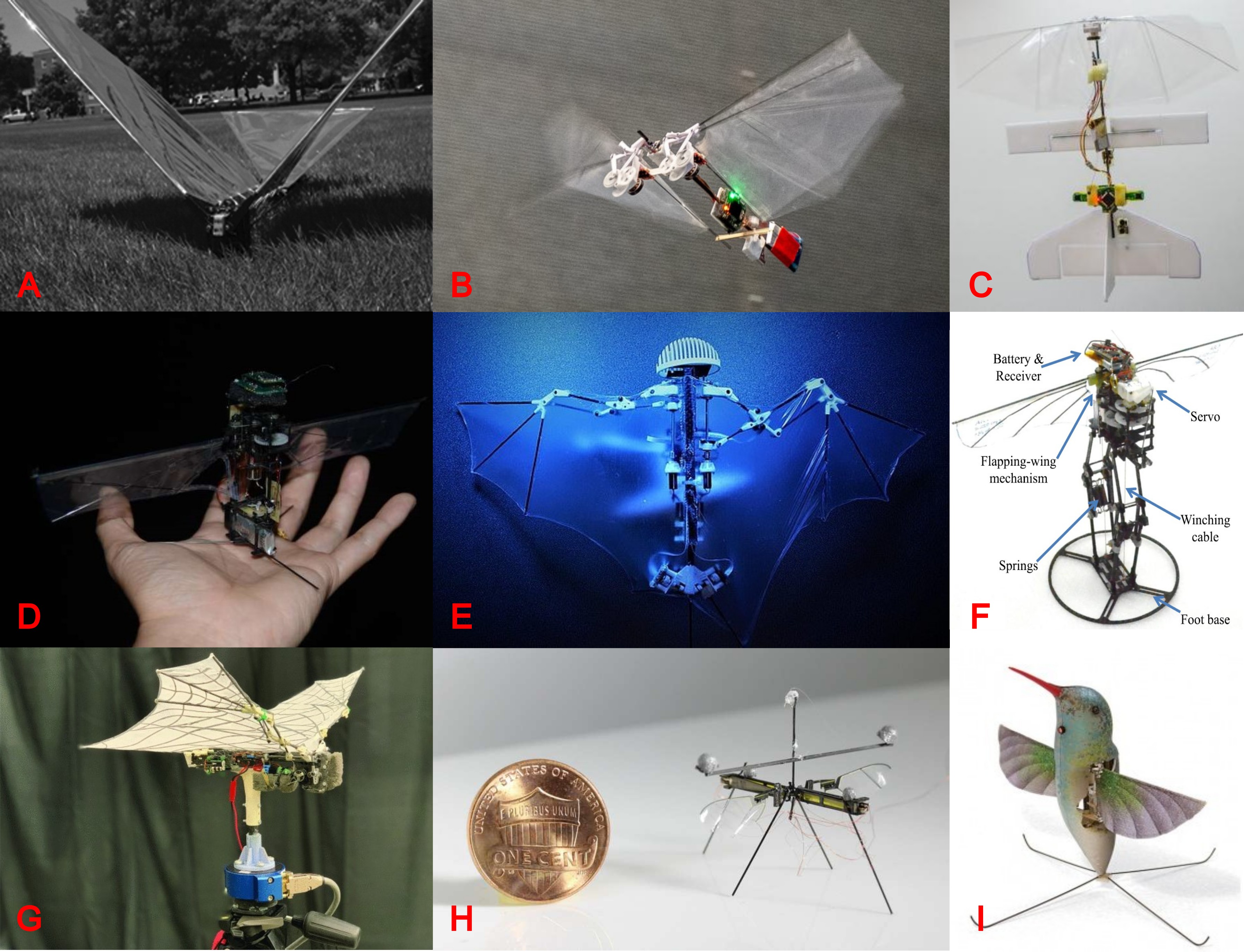}
    \caption{Illustrates state-of-the-art bio-inspired aerial robots. }
    (A) Robo Raven from \cite{gerdes2014robo}. (B) DelFly Nimble from \cite{karasek2018tailless}. (C) DelFly Explorer from \cite{de2014autonomous}. (D) KUBeetle from \cite{phan2017design}. (E) Bat bot from \cite{ramezani_biomimetic_2017}. (F) Jump-flapper from \cite{truong2019design}. (G) Bat Bot 2.0 from \cite{hoff2021bat}. (H) RoboFly from \cite{chukewad2021robofly}. (I) Nano-Hummingbird from \cite{keennon2012development}.
    \label{fig:litreview}
\end{figure}

Present-day drones, encompassing both rotary and fixed-wing types, face significant challenges navigating confined spaces. Although fixed and rotary-wing platforms are frequently employed for surveillance and reconnaissance, there is a burgeoning interest in equipping these micro aerial vehicles (MAVs) with advanced sensor suites. This enhancement aims to leverage their aerial mobility to detect and monitor hazardous conditions in residential environments. Notably, while platforms like quadrotors exhibit agile movement and commendable fault tolerance in hostile settings, their susceptibility to collisions stems from their rigid construction. Recent design approaches have begun to integrate soft and flexible materials into these systems. However, aerodynamic efficiency considerations limit the extent to which rotor blades or propellers can be fabricated from highly flexible materials.

The flight mechanisms of bats provide significant guidance for pioneering MAV designs suitable for residential operations. The bats' distinct body articulation, or morphing ability, underpins their unmatched aerial prowess. By contracting their wing area during upstrokes and expanding it during downstrokes, bats optimize lift generation\cite{tobalske_biomechanics_2000}. Some bat species are recognized for employing differential inertial forces to execute agile zero-angular momentum turns\cite{riskin_upstroke_2012}. Further, biological studies indicate that the intricate musculoskeletal systems of these animals offer enhanced impact absorption, potentially increasing their resilience during collisions.

\begin{figure}
    \centering
    \includegraphics[width=1.0\linewidth]{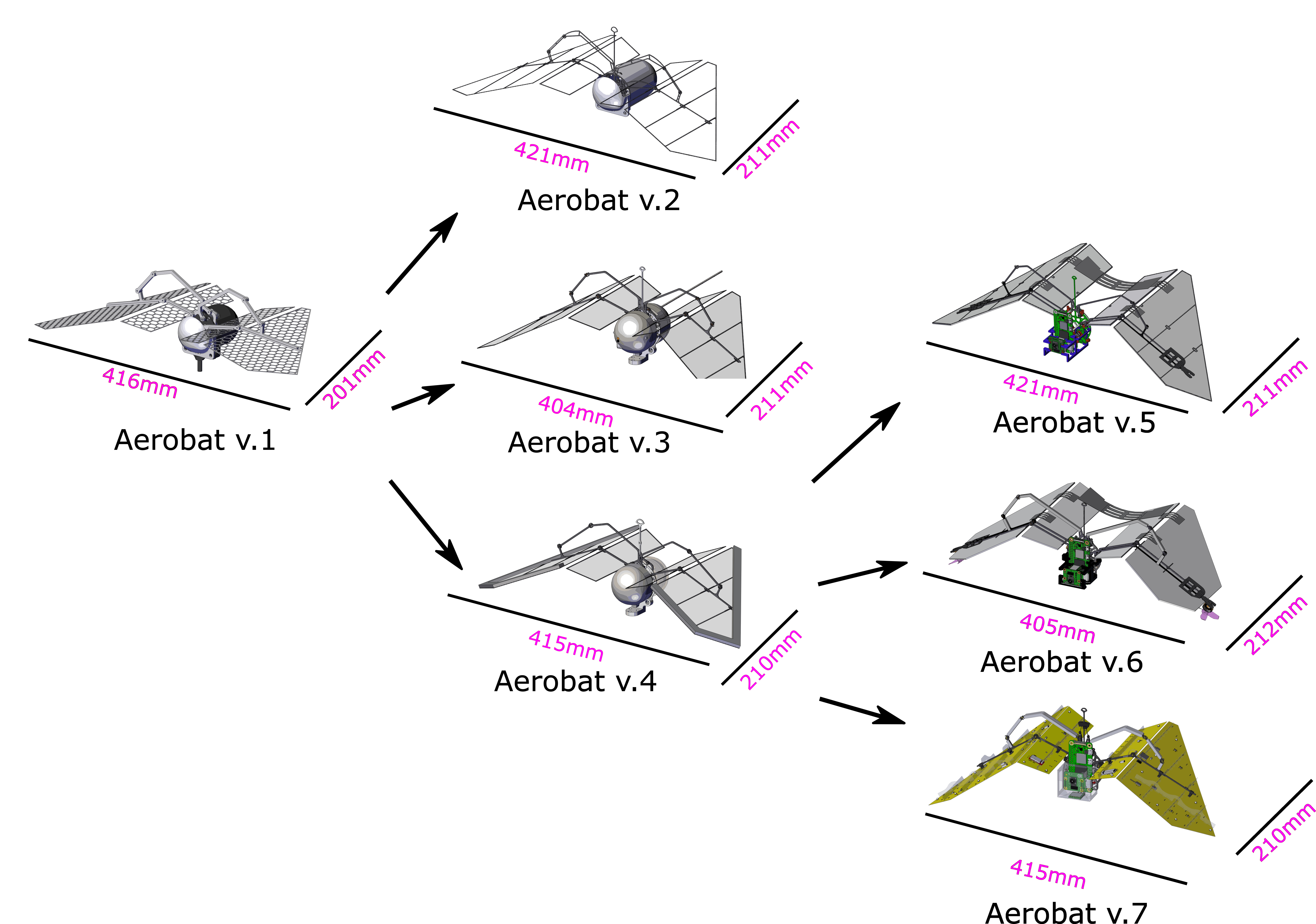}
    \caption{Illustrates Aerobat's iterative designs and how this Thesis contributes to these evolutionary designs.}
    \label{fig:all_aerobat}
\end{figure}

Specifically, this Master's thesis aims to expand upon previous work (as depicted in Fig. \ref{fig:all_aerobat}) by integrating polyimide flexible PCB wings into the Aerobat, accompanied by a series of structural optimizations for the membrane wing and guards. The innovative integration of a flexible PCB into the wing membrane of a bat-inspired robot offers transformative capabilities. Not only does this integration serve a dual role as both a structural component and an electronic platform, but it also introduces several distinct advantages. Key among these is the dynamic morphing ability, where the PCB's flexibility allows the wing to adapt shape during flight. This adaptability can be enhanced by direct sensor integration, capturing real-time data on wing deformations, aerodynamic interactions, and temperature variations. The merged roles of structural and electronic components in the flexible PCB offer a notable reduction in the robot's weight and a more compact assembly, eliminating extraneous wiring and connectors. Integrated communication modules on the PCB provide efficient data transmission without additional weight. This streamlined integration ensures robustness during flight, with the PCB's inherent resilience to bending and flexing. In-flight adaptations are further bolstered by real-time data processing capabilities, facilitated by onboard microcontrollers. The PCB's design also allows for the potential integration of energy-harvesting elements, converting wing movements into electrical power. Heat dissipation, crucial given the potential heat generation by wing actuators, is enhanced by the PCB's material choice. Through a careful balance between flexibility and rigidity, the flexible PCB has the potential to redefine the operational and functional horizons of bio-inspired robots.

Bat wings, with their membranous structures, offer exceptional functionalities \cite{tanaka_flexible_2015} that serve as an archetype for advancing the design of drones intended for confined environments such as pipes, narrow corridors, tunnels, and HVAC channels. Diverging from other avian species, bats boast a uniquely articulated musculoskeletal system. This system not only augments their resilience against impacts but also equips them with adaptive and versatile locomotion behaviors \cite{riskin_quantifying_2008}\cite{wang2023legged}\cite{salagame2022letter}. Such traits help circumvent the creation of continuous, forceful air jets—a primary obstacle that hinders multi-rotors from operating in tight spaces. Rather than exerting a continuous force on the air, bats leverage their structural flexibility to modulate force distribution across their wing membranes through intermittent jet generation. The synergy of wing pliability, sophisticated kinematics, and rapid muscle actuation empowers bats to swiftly adjust their body configuration in mere tens of milliseconds, which underscores their unmatched flight manipulation capabilities \cite{azuma_akira_biokinetics_2006}.


\section{Thesis Focus}
\label{sec:focus}
Several attempts have been made to copy the flapping flight of animals, including insects, birds, bats, etc., ranging from smaller insect-sized robots or MAVs \cite{phan_insect-inspired_2019, farrell_helbling_review_2018, ma_controlled_2013, chukewad_robofly_2020, tu_untethered_2020, rosen_development_2016}, to bat or small bird-sized robots with a wingspan between 20 and 60 cm \cite{hoff_synergistic_2016, ramezani_bat_2016, ramezani_biomimetic_2017, hoff_optimizing_2018, sihite_computational_2020, de_croon_design_2009, peterson_wing-assisted_2011, wissa_free_2015}, and larger robots with wingspan larger than 1 m \cite{send2012artificial, gerdes2014robo}. Unfortunately, most of these examples fail to copy dynamic morphing capabilities manifested by the powered flight of animals. Almost all of these examples are incapable of hovering while carrying computers and cameras. 
The primary aim of this thesis is to augment the Aerobat's iterative design by integrating a guard mechanism and motorized wings, ensuring its safe operation within the RISE cage. While flapping-wing MAVs are not novel—bird-inspired design paradigms for such vehicles trace back to the early 20th century—the impetus for their development often arises from biomimicry, the emulation of natural systems and structures. What sets the Aerobat apart is its distinctive tailless design combined with a morphing wing structure\cite{sihite2022unsteady}. This morphing capability allows the wing to adapt its shape in alignment with the flapping motion's phase, optimizing efficiency and attenuating its fluidic signature, a crucial feature for operation within constrained environments like RISE.

The dynamic nature of the Aerobat's morphing wings presents formidable challenges to its flight control \cite{ramezani_lagrangian_2015, mangan_describing_2017, mangan_reducing_2017, hoff_trajectory_2019, sihite_enforcing_2020}. Previous studies on the Aerobat primarily centered on creating simulation models for control strategy exploration \cite{sihite_enforcing_2020} and the incorporation of embodied locomotion through morphological adaptations utilizing mechanical intelligence \cite{sihite_integrated_2021}. The aircraft's tailless configuration further complicates its flight control dynamics. As the Aerobat depends solely on its wing structure for flight control and stability, devoid of a tail, it is imperative to address the inherent instabilities stemming from both the morphing wings and its tailless design, thereby facilitating the development of a robust flight control system.

\begin{figure}
    \centering
    \includegraphics[width=0.8\linewidth]{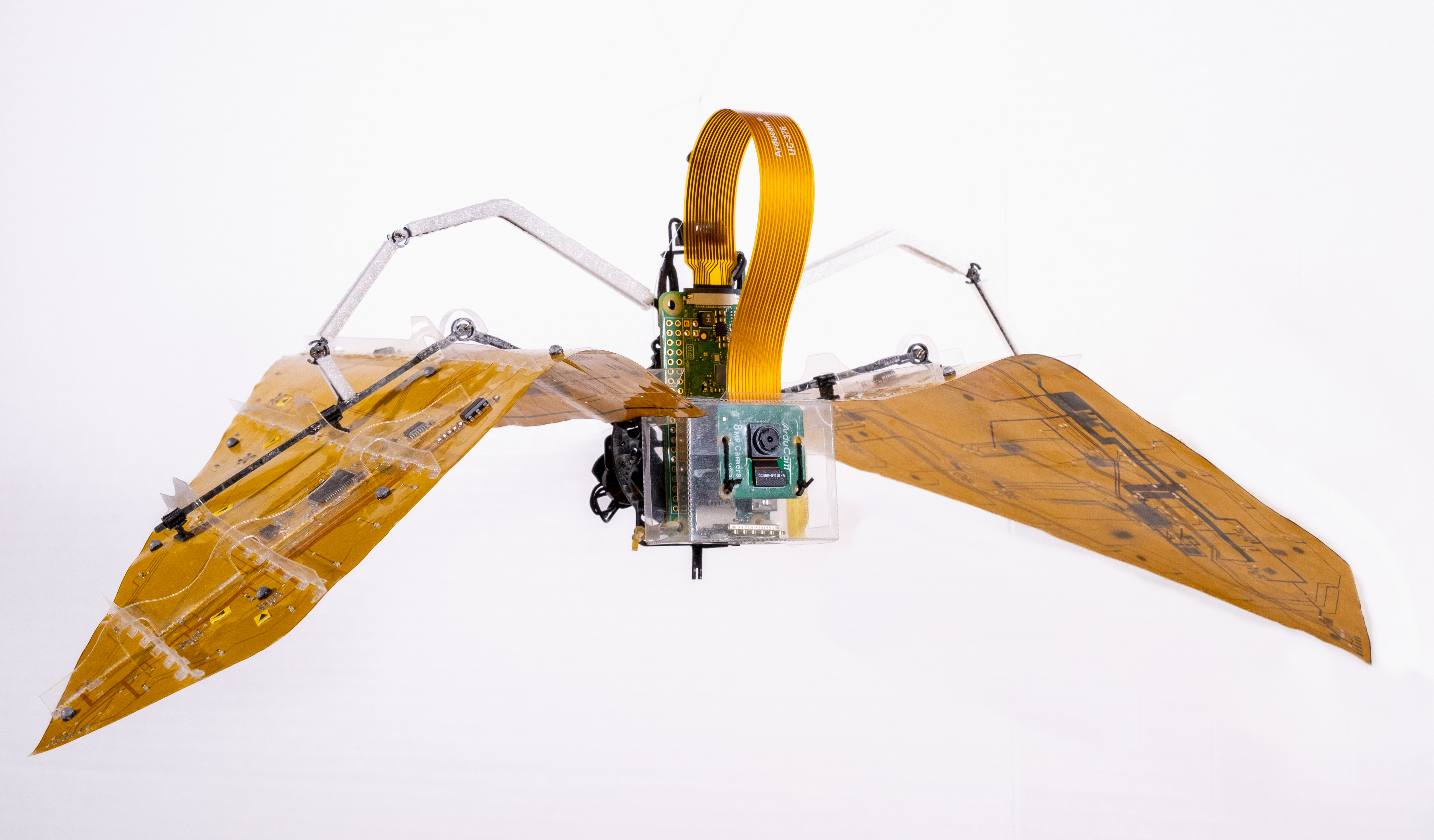}
    \caption{Illustrates Aerobat's prototype ($\gamma$ prototype) created as part of this MS Thesis to study flight in RISE cage (Fig.~\ref{fig:rise}).}
    \label{fig:aerobat-gamma}
\end{figure}

To achieve this, this MS Thesis envisions a protective device that mitigates the instability caused by wing folding and provides flight control assistance. The protective device, the guard, balances the unstable dynamics of Aerobat. 
We designed the wing membrane using flexible PCB, shown in (Fig.~\ref{fig:aerobat-gamma}), where we attach sensors, motor drivers, communication modules, and microcontroller unit. The components attached on the wings will be used as apart of our future work to incorporate mechanical intelligence to facilitate control in this robot. Moreover, the simplification of the wing bones has been undertaken. The wing's skeletal framework and electronic cage have been reconfigured utilizing origami techniques and living hinges. This dual approach not only reduces the overall weight of the Aerobat but also bolsters its structural integrity, thus augmenting both agility and stability. The combination of Guard and structural optimization empower the Aerobat to achieve fully self-contained, secure hovering flights within the confines of the RISE cage.
To conclude, the Thesis contributions can be divided into two aspects:
\begin{itemize}
    \item We design a novel preload mechanism to address the slack issue of the Aerobat arm during the terminal phase of the downstroke, a pivotal factor in enhancing the lift of the Aerobat's flapping wing. Simultaneously, optimization of the wing's skeletal structure and the implementation of a sophisticated living hinge design are undertaken to counteract the cascade buckling phenomenon in the membrane wing. Additionally, employing origami technique, an ultra-light electronic cage is engineered to accommodate the microcontroller, IMU, and camera, facilitating precise attitude control of the Aerobat. 
    \item We design and prototype a light-weight guard to aid the Aerobat platform in flight, which will be vital in the safe operations of Aerobat inside very tight spaces. As part of the guard design, the Thesis looks into component selection, mechanical design, modeling, and flight control design for the guard.
\end{itemize} 

\section{Thesis Structure}
\label{sec:structure}
The structure of this thesis is designed to elucidate our research comprehensively. We introduce the \textit{Aerobat}, a platform conceived at Northeastern University, elucidating its fundamental design tenets. Subsequent sections engage in the dynamic modeling of the Aerobat, concentrating on two main components: (1)the wing structure optimization, which plays a crucial role in the platform's performance, and (2) the Guard, a protective structure that also functions as an auxiliary for flight. In the subsequent section, we discuss the intelligent design details of the wing structure and electronic cage. We also discuss the design and control methods of the Guard, elaborating on how these methods contribute to the platform's stability and efficiency. Finally, the paper concludes with evaluating our findings, lessons learned, and future work, offering insights into potential improvements and new directions for further investigation.

\chapter{Hardware Design of \textbf{\textit{Aerobat} $\gamma$}}

This chapter gives an overview of mechanical design, kinematics, and material selections carried out in designing Aerobat $\gamma$. The overall computational structures of the wings, transmission mechanism design, and electronics are inherited from previous revisions of Aerobat and have not been designed as part of this MS thesis. However, for the sake of the completeness of this Thesis, a brief overview of computational structures will be provided.  

\section{Computational Structures}
\label{sec:computational-structure}
NU's Aerobat is a tail-less flapping wing robot that unlike existing examples are capable of significantly morphing wing structure dynamically during each gait cycle which is a fraction of a second. This robot, which weighs roughly 40-50 grams depending on the onboard sensors, with a wingspan of approximately 30 cm, was developed to study the flapping-wing flight of bats. 

Aerobat utilizes a computational structure, called the \textit{Kinetic Sculpture} (KS) \cite{sihite_computational_2020}, that introduces computational resources for wing morphing. The KS is designed to actuate the robot's wings as it is split into two wing segments: the proximal and distal wings, which are actuated by the equivalent of shoulder and elbow joints. 
The KS captures two crucial modes within the bat flapping gait, specifically the plunging and elbow flexion/extension modes. These modes collectively emulate the dynamic wing morphing observed in bat flight, where the wing folds and expands during the upstroke and downstroke, respectively. This folding action mitigates negative lift generation while concurrently reducing wing inertia, facilitating swifter transformation during the upstroke. This inherent morphing capability contributes to the attainment of a highly efficient flapping gait, aligning with the objective pursued in our design. Aerobat can flap at a frequency of up to 8 Hz using its onboard electronics.

The components and main body of the Aerobat are made of laser-cut carbon fiber plates assembled using copper rivets. Consequently, the Aerobat inherits the characteristics of rigid carbon materials, making it resistant to deformation when subjected to external forces. 

\subsection{Embodiment and Dynamic Morphing Wing Flight}

An untethered, self-sustained, and autonomous robotic platform that can mimic bird and bat explosive wing articulations is a significant design problem after noting the prohibitive design restrictions such as payload, size, power, etc. Embodiment, which means using mechanical design and computational structures to subsume the responsibility of closed-loop feedback to save power, payload, and computational resources, is a powerful concept to consider to address these challenges. However, the embodiment of aerial locomotion has remained unexplored.

Micro Aerial Vehicles (MAVs) showcasing morphing bodies \cite{chang2020soft,chin_inspiration_2017,tang_autonomous_2018} have set themselves apart due to their superior performance. Yet, they remain less explored compared to the myriad of conventional flapping-wing robots that span a spectrum from insect-inspired MAVs \cite{farrell_helbling_review_2018} to larger avian-inspired counterparts \cite{di2020bioinspired,karasek2018tailless,roberts2017modeling, holness_design_2019}.

Typically, flapping MAVs possess single-segment wings, articulated either to flap about a fixed axis \cite{madangopal_biologically_2005,yang_dove_2018} or a rotating one, adjusting the wing's angle of attack \cite{sane_lift_2001,send2012artificial}. Such designs often neglect wing folding. The supination-pronation motion, crucial for generating lift during upstrokes, is either passively executed or to a degree doesn't mirror biological entities.

Recognizing the significance of wing folding in avian locomotion, some recent designs have emerged \cite{send2012artificial}. Yet, stringent design challenges often hinder replicating the pronounced mediolateral or flexion-extension motions evident in natural flight mechanisms. For instance, the \textit{SmartBird} \cite{send2012artificial} has dual-segment wings that flex at the elbow joint, optimizing lift generation during the motion cycle.

Other works\cite{ramezani_bat_2016,hoff_synergistic_2016,hoff_optimizing_2018,mangan_reducing_2017}  attempted to design armwing retraction mechanisms and used the opportunity to study the underlying control mechanisms \cite{ramezani2017describing,ramezani2015lagrangian,ramezani2016nonlinear,hoff_trajectory_2019} based on which bats perform sharp banking turns and diving maneuvers. The morphing wing design introduced by \cite{hoff_optimizing_2018} considered substantially fewer joints in an untethered system by erecting a \textit{kinetic sculpture} that embodied several biologically meaningful modes from bats. Contrary to \cite{hoff_optimizing_2018}, many string-and-pulley-activated joints were incorporated in the morphing wings introduced by \cite{bahlman_design_2013} and \cite{colorado_biomechanics_2012} that allowed a greater control authority over independent joint movements. However, these designs were tethered. All of these morphing wings have achieved great success in copying the kinematics and dimensional complexity of bat flight apparatus. However, the armwing mechanisms present in these robots could not copy the dynamically versatile wing conformations found in bats.

\subsection{Aerobat's Design Philosophy Based on Coalescing Mechanical Intelligence and Closed-Loop Feedback}

However, the emerging ideas surrounding achieving computation in robots through sophisticated interactions of morphology have begun to change motion design and control in robots with prohibitive design restrictions. Such a computation called \textit{morphological computation} or \textit{mechanical intelligence}\cite{hauser_role_2012}, draws our attention to the obvious fact that there is a common interconnection -- and in some morphologies these couplings are very tight -- between the boundaries of morphology and closed-loop feedback. 

Traditionally, controllers function within abstract computational domains, either embedded within computational layers or explicitly programmed. However, when mechanical interactions are harnessed to facilitate computation, morphology can actively participate in the system's computational processes. Consequently, a well-conceived mechanical design can ease control demands by assuming part of the computational burden. Incorporating such biologically-inspired computational structures can be advantageous for the design of morphing MAVs. Historically, these structures were neglected, largely due to complexities in design and fabrication.

\section{Mechanical Design of Wing-Arms in \textbf{\textit{Aerobat $\gamma$}}}
\label{sec:design}

\begin{figure}[t]
    \centering
    \includegraphics[width=1\linewidth]{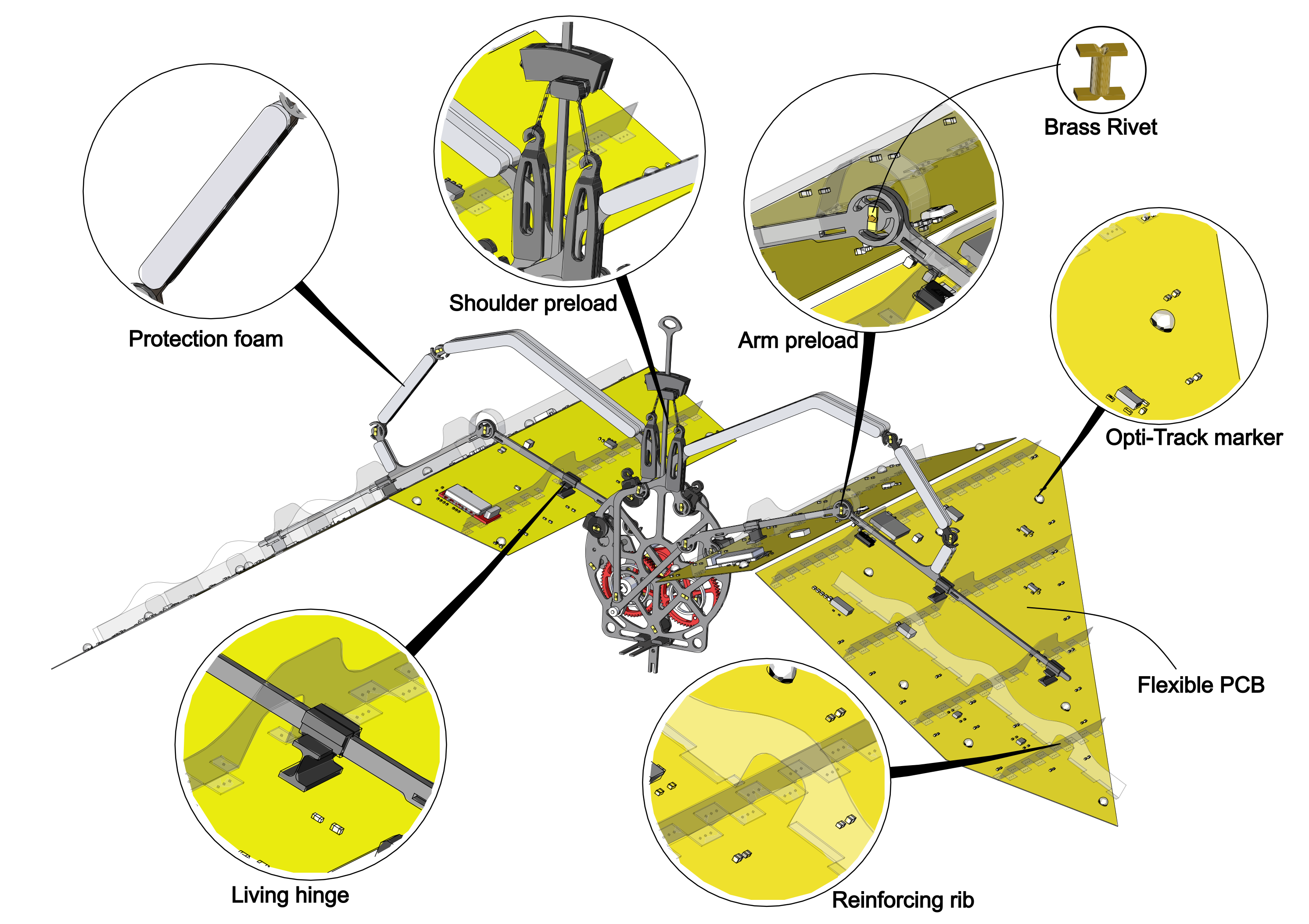}
    \caption{Illustrates the details of wing design }
    \label{fig: Gamma wing design details}
\end{figure}

In order to successfully design an armwing structure that can mimic the speed and flexibility of a natural bat wing, we must work to achieve a set of design criteria: 
\begin{itemize}
    \item (a) a mechanical structure that mimics as many meaningful degrees of freedom (DoF) as possible from the natural bat wing, 
    \item (b) a robust and flexible wing structure that facilitates control through morphological computation, and 
    \item (c) a small, lightweight, and compact mechanism.
\end{itemize}

The meaningful DoFs we considered in this paper is the plunging motion along with the wing extension/retraction, where the control is facilitated by either changing the wing morphology or by directly articulating the armwing kinetic sculpture. As illustrated in Fig.~\ref{fig:wing_linkage}, a single motor powers the entire flapping mechanism, actuating the KS to execute the robot's flapping gait. The KS enables the incorporation of elbow flexion and extension, a critical mode in bat flight, allowing the wing to fold during the upstroke and minimize negative lift.



To counteract the slackness in the humerus and radius, two distinct pre-loads, one for the shoulders and another for the elbows, are implemented, as illustrated in Fig. \ref{fig: Gamma wing design details}. The combination of material strength and weight constraints leads to unexpected slackness in both the shoulders and elbows during the downstroke's conclusion. The application of these pre-loads significantly mitigates the slackness in the entire arm skeleton at the downstroke's termination, thereby substantially augmenting the Aerobat's flapping ability and resulting in increased lift generation. The shoulder pre-loads are fabricated from micro carbon fiber filled nylon(Onyx) which is 3d printed using Markforged Mark Two 3D printer. This machine uses a continuous fiber reinforcement process to produce strong parts with 100$\mu$m Z layer resolution reliably. 

\subsection{Wing Membrane Mechanism Design}
During wing retraction, the elbow joint penetrates the wing membrane, resulting in irreversible damage to the flexible PCB. To mitigate this damage, a strategic decision was made to create a safe zone between the wing membrane and the wing arm. However, this approach may lead to the buckling of laminated structures effect to some extent. Specifically, when the wing arm and wing membrane are fixed at the shoulder joint and wingtip, simultaneous inward bending of the arm and membrane can reduce the inner membrane's installation space below the arm length, subjecting the membrane to compressive stress. Conversely, simultaneous outward bending of the arm and membrane can increase the outer membrane's installation space beyond the inner arm's length, resulting in tensile stress on the membrane. To address the potential membrane damage or wing resistance due to the buckling of laminated structures effect, an innovative living hinge structure has been meticulously designed.

\begin{figure}
    \centering
    \includegraphics[width=0.7\linewidth]{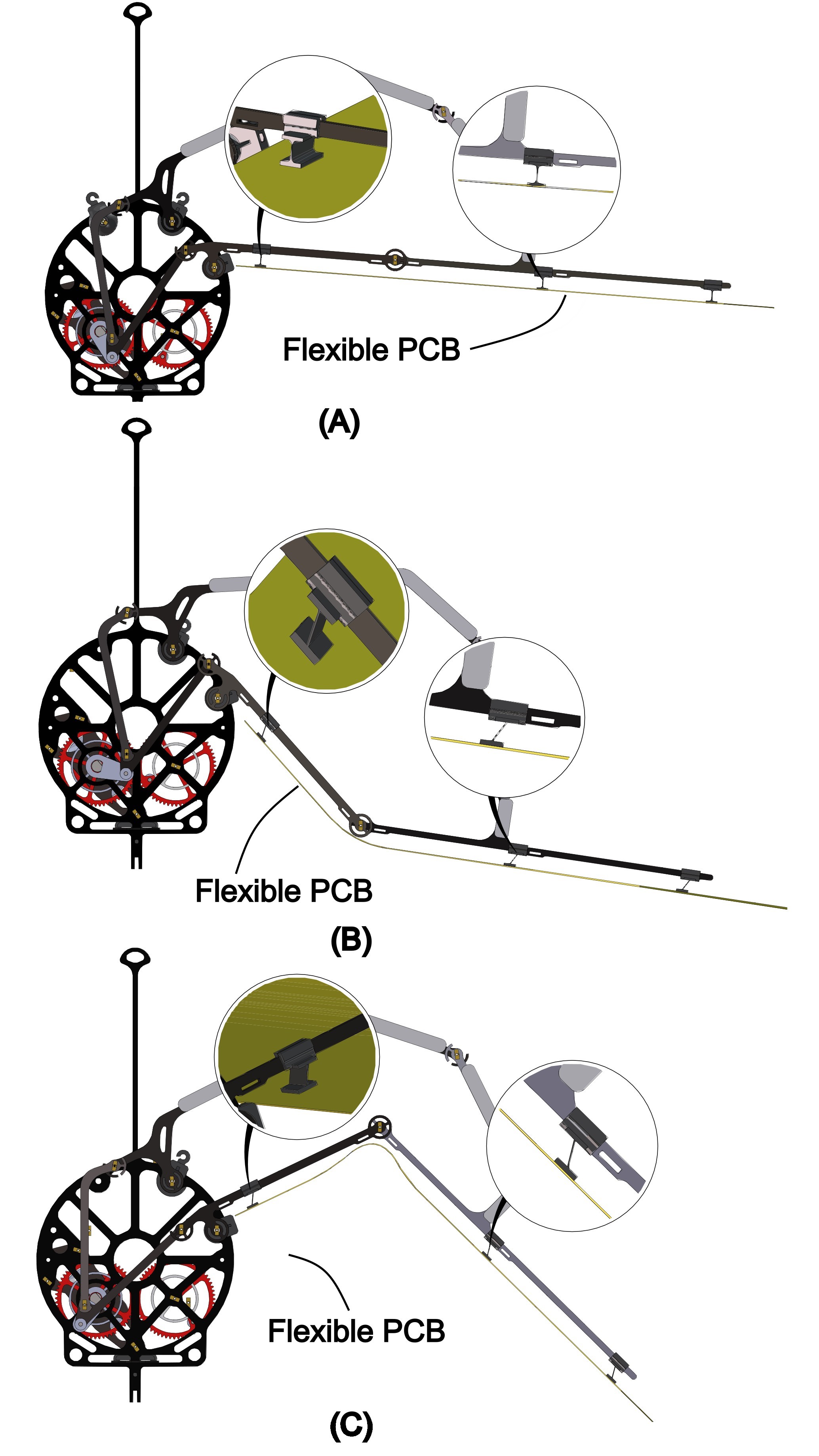}
    \caption{\textbf{Living hinge deformation in three configurations.} (A) extension configuration. (B) outer retraction configuration. (C) inner retraction configuration.}
    \label{fig: flexible pcb 3 configration}
\end{figure}

The Living hinge structure, fabricated in an I-shape using Onyx material on a Markforged Mark two 3D printer, exhibits thicker upper and lower ends employed for bonding the arm and flexible PCB, respectively. The central area is designed as an extremely thin and highly flexible sheet, allowing the living hinge to bend in the direction requiring compensation. The wing incorporates the living hinge in the extension configuration, depicted in Fig.~\ref{fig: flexible pcb 3 configration}A, representing its natural state. When the Flexible PCB is bent, it positions itself on the outer side of the arm (as shown in Fig.~\ref{fig: flexible pcb 3 configration}B). To ensure the flexible PCB and the arm bend at the same angle, the living hinge tilts towards the elbow joint, compensating for the buckling effect of laminated structures. Conversely, when the Flexible PCB bends to the inner side of the arm (Fig.~\ref{fig: flexible pcb 3 configration}C), the living hinge inclines away from the elbow joint to maintain the desired angle. The living hinge plays a pivotal role in reducing the drag of the Aerobat flapping wing, safeguarding the integrity of the membrane, and enhancing the performance of the Aerobat flapping wing.

\renewcommand\floatpagefraction{.9}
\renewcommand\topfraction{.9}
\renewcommand\bottomfraction{.9}
\renewcommand\textfraction{.1}
\setcounter{totalnumber}{50}
\setcounter{topnumber}{50}
\setcounter{bottomnumber}{50}

\subsection{Electronic Cage Design}

\begin{figure}
    \centering
    \includegraphics[width=1\linewidth]{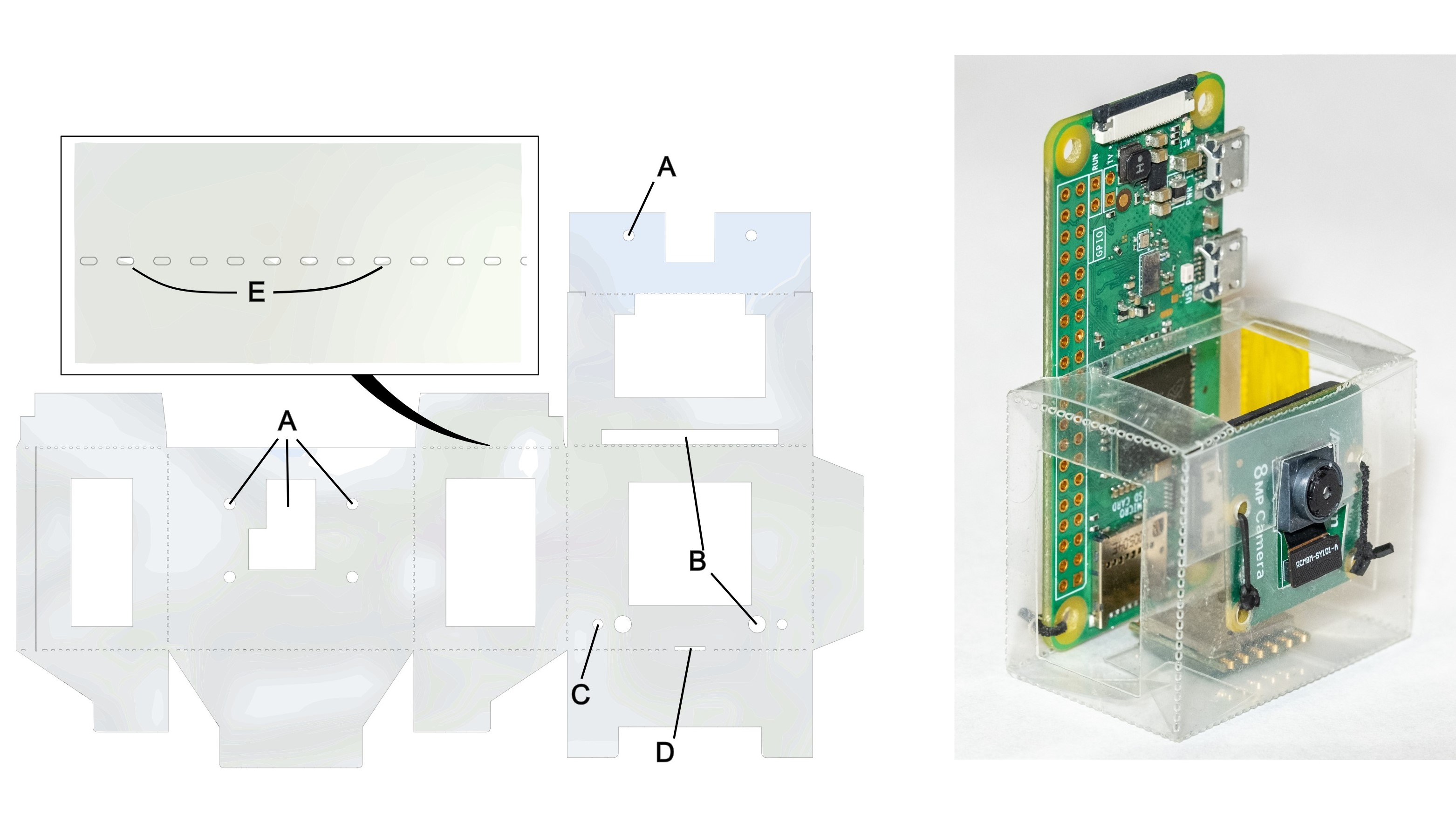}
    \caption{\textbf{Aerobat $\gamma$ electronic cage.} (A) camera mounting holes. (B) PCB mounting holes. (C) rubber band mounting holes. (D) Aerobat mounting holes. (E) fold lines.}
    \label{fig: gamma electronic cage}
\end{figure}

Over the past decade, origami has found numerous novel applications in various areas of robotics. A novel electronic cage design is proposed in this research, featuring a lightweight, economical, and vandal-resistant structure. The design is compared to a conventional PLA electronic cage. The electronic cage is designed using origami techniques, manually folded from a 0.2-mm-thick polypropylene sheet as shown in Fig.~\ref{fig: gamma electronic cage}. To facilitate folding and ensure accurate geometric replication, elliptic perforations were engraved along the fold lines and mounting holes for electronic components using a CO$_2$ laser cutter(Universal Laser Systems VLS3.75). The folded polypropylene sheet forms a cuboid structure that houses a Raspberry Pi Camera Module 2, WITMOTION WT901 9-Axis IMU, and Raspberry Pi Zero 2 W camera with rubber, securely fixed on the Aerobat's base plate.

Compared to the PLA material design produced by the Ultimaker Cura 3D printer, the polypropylene-origami design is significantly lighter in weight, reducing from 7.2g to 1.7g, representing a weight reduction of approximately 76.4$\%$. Additionally, polypropylene's excellent mechanical toughness and elasticity enable better impact force dissipation and enhanced protection of electronic components. Furthermore, the specific molecular arrangement and structure of polypropylene make it resistant to lateral bending, thereby ensuring the overall strength of the electronic cage.

\subsection{Transmission Mechanism}

\begin{figure}[htp]
    \centering
    \includegraphics[width=1\linewidth]{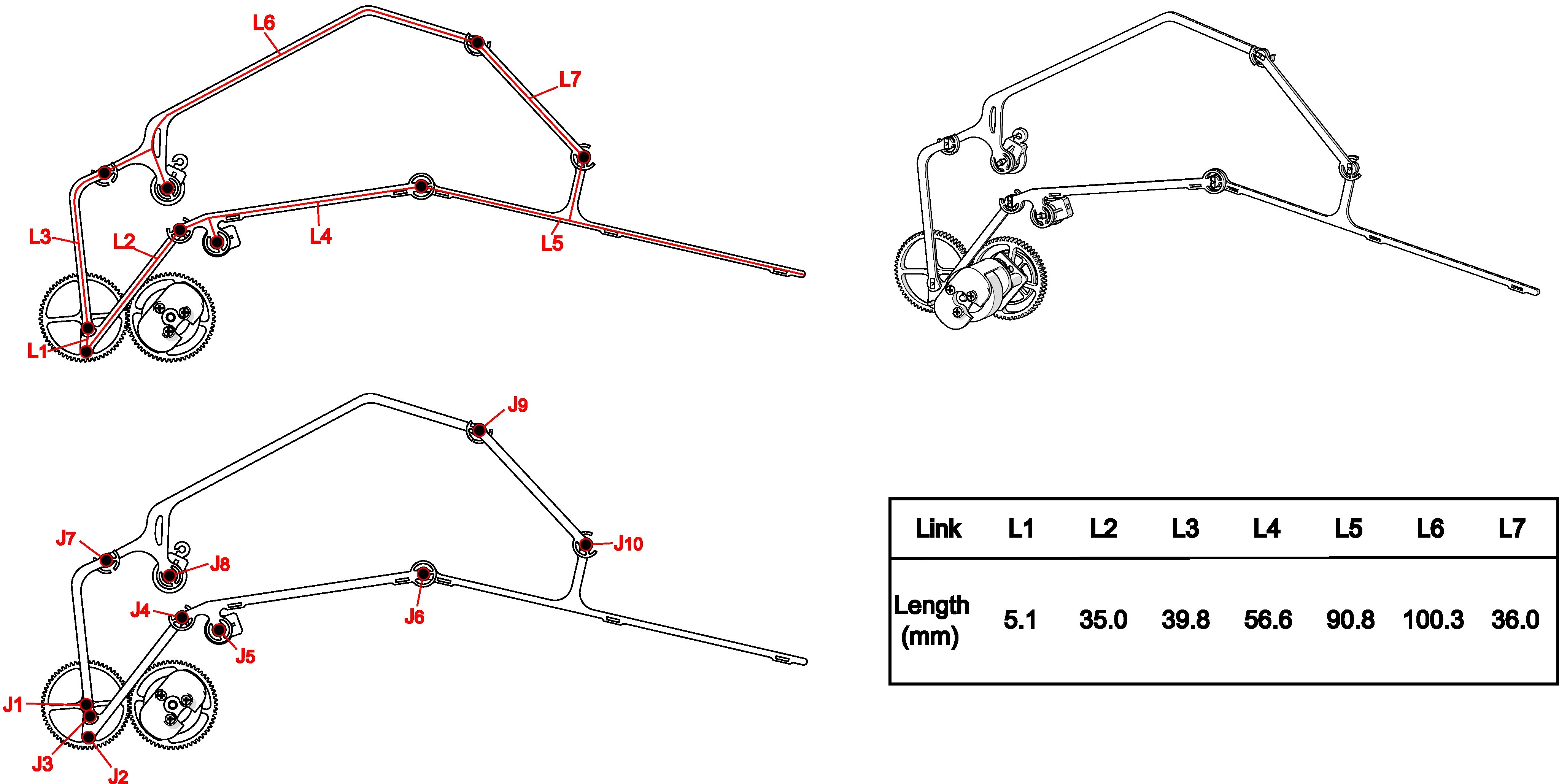}
    \caption{The links and joints present in the wing linkage mechanism, which is composed of 7 links ($L_i, i \in \{1,\dots, 7\}$) and 10 joints ($J_k, k \in \{1,\dots, 10\}$). The crank mechanisms drive links $L_2$ and $L_3$ which actuate the shoulder and elbow joints ($J_4$, $J_5$, $J_7$ and $J_8$) respectively.}
    \label{fig:wing_linkage}
\end{figure}

The armwing mechanisms are driven by two sets of crank and four-bar mechanisms, as illustrated in Fig. \ref{fig:wing_linkage}. This mechanism articulates the linkages representing the humerus and radius bones found in a bat's arm. Both crank mechanisms operate at the same frequency but with a distinct phase difference, $\Delta \phi$, which dictates the desired wing extension and retraction during specific moments within a wingbeat. The wing structure consists of four links and seven joints per wing, to which an additional three links and three revolute joints are incorporated through the gears and crank mechanism, resulting in a total of seven links and ten joints per wing. Joint 1 denotes the gear or motor angle responsible for driving the linkages, thus governing the flapping gait. Joints 5 and 6 represent the wing shoulder and elbow joints, respectively. The wings are connected to linkages 4 and 5, representing the proximal and distal wings.

The humerus and radius links are 50 mm and 90 mm respectively. Due to space constraints, the four-bar mechanisms are placed off-plane and parallel to each other.~
To achieve a symmetric wing assembly in the Aerobat, the drive and driven gears responsible for driving the four-bar mechanism must be aligned parallel to the central axis. The Happymodel EX1102 9000KV brushless motor is positioned behind the right gear and connected to the drive gear via a Pololu gearbox with a reduction ratio of 75:1, facilitating the transmission of rotational motion to the gear. In future untethered experiments, we plan to place the battery behind the driven gear to counterbalance the weight of the gearbox and motor, thereby enhancing the Aerobat's flight performance. This way, each side of the wing can utilize the same wing structure and the mechanisms can be connected using a spur gear or other means of power transmission. This configuration results in a horizontally-symmetric but off-plane wing skeletal structure. However, this should not be a major issue because the wing membranes can be attached in a symmetric fashion. 

\subsection{Computational Structure Output}
\label{ssec:comp-output}

\begin{figure}[t]
    \centering
    \subfloat[Unoptimized armwing angles.]{%
    \label{fig:plot_wing_angles_a}\includegraphics[clip,width=\columnwidth]{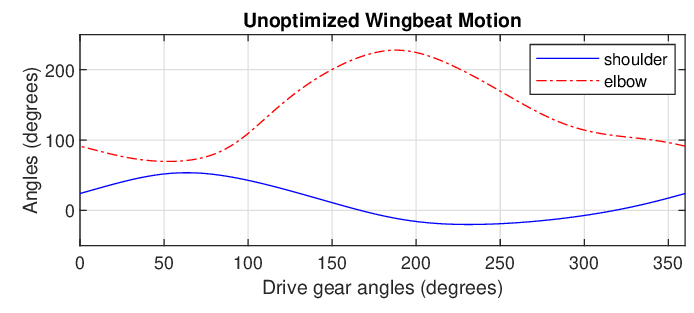}%
    }
    
    \subfloat[Optimized armwing angles and their target trajectories.]{%
    \label{fig:plot_wing_angles_b}\includegraphics[clip,width=\columnwidth]{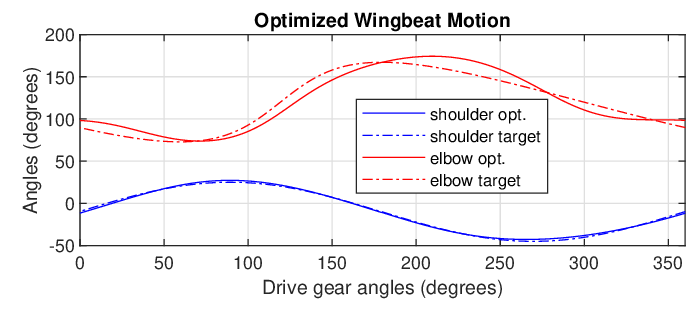}%
    }
    
\caption{The unoptimized and optimized armwing shoulder and elbow angles within a single wingbeat. The optimization has successfully found the parameters that result in close tracking of the target trajectory.}
\label{fig:plot_wing_angles}
\end{figure}

The ideal flapping motion we are looking for has the following properties: (1) the wing extends and retracts during downstroke and upstroke respectively, (2) the wing is already partially expanded before the downstroke motion begins. The desired trajectories $\hat{\theta}_s$ and $\hat{\theta}_e$ can be seen in Fig. \ref{fig:plot_wing_angles_b}, which are defined as the following sinusoidal functions
\begin{equation}
\begin{gathered}
\hat \theta_s = 35^\circ\, \sin(\phi) - 10^\circ\\ 
\hat \theta_e = -0.5\,\tan^{-1}\left( \tfrac{-0.5\,\sin(\phi + 2\pi/3)}{1 + 0.5\,\cos(\phi+2\pi/3)}  \right) \, 45^\circ + 120^\circ,
\end{gathered}
\end{equation}
where $\phi \in [0,2\pi)$. $\hat{\theta}_e$ is a skewed sinusoidal function that allows the wing to expand faster than the retraction and have a full wingspan in the middle of the downstroke.

Figure \ref{fig:plot_wing_angles} shows the shoulder and elbow angles simulation for one flapping cycle using the rigid body kinematic. The unoptimized design trajectories, shown in Fig. \ref{fig:plot_wing_angles_a}, has a range of $\theta_s \in [ -20^\circ, 54^\circ]$ and $\theta_e \in [ 70^\circ, 228^\circ]$. However, the maximum elbow angle of $228^\circ$ raises some concerns since that indicates a nontrivial joint hyperextension which might have an adverse effect on the armwing structure.

The design optimization process effectively identified the design parameters, $\bm{q}$ listed in Table \ref{fig:wing_linkage}, which have 15.8\% average difference compared to the initial values and closely follow the target trajectories as shown in Fig. \ref{fig:plot_wing_angles}, with $R^2$ values of 0.997 and 0.920 for $\theta_s$ and $\theta_e$ respectively which indicate a good fit. The optimized trajectory demonstrates motion ranges of $\theta_s \in [-43^\circ, 27^\circ]$ and elbow angle range of $\theta_e \in [74^\circ, 174^\circ]$, notably avoiding the joint hyperextension observed in unoptimized design. The elbow joint and wingtip trajectories of the optimized design can be seen in Fig. \ref{fig:plot_wing_angles}. Additionally, it's vital to ensure that hinge bending angles don't exceed safe thresholds, potentially leading to breakage. Specifically, Joints 6, 9, and 10—all connected to the radius link $L_{4}$ —exhibit maximum bending angles ranging between $80^\circ$ and $90^\circ$ that bend significantly more towards one side than the other. Joint 4 has a maximum bending angle of $65^\circ$, while the other joints have a maximum bending angle of less than $45^\circ$.

\chapter{Aerobat \textit{Guard} Design}

The primary roles of the Guard system include:
\begin{itemize}
    \item protecting Aerobat,
    \item help stabilization of roll, pitch, and yaw angles
\end{itemize}
The development of an efficient protective structure for Aerobat requires a nuanced equilibrium among strength, weight, and aerodynamics. This equilibrium is pursued through an iterative design methodology, meticulously addressing key factors at every iteration. The process systematically evaluates strength, payload, and aerodynamic interference to discern the optimal solution. This section elucidates this methodology, underscoring the significance of material choice, structural integrity, impact resilience, aerodynamic factors, and weight management in fashioning a resilient and efficient guard.

\section{Frame Design}


The designed deformable elastic structure, with its ellipsoid shape, aims to offer comprehensive protection to the Aerobat, taking into account the range of motion of its wings. The major axis of the ellipse is strategically longer than the wingspan, ensuring optimal coverage and providing inherent structural integrity with a uniform stress distribution. This design effectively redistributes impact stresses and prevents their transfer to the Aerobat by utilizing its lightweight carbon-fiber structure to dampen the effects of collisions or crashes with the environment.

\begin{figure}[htp]
    \centering
    \includegraphics[width=1\linewidth]{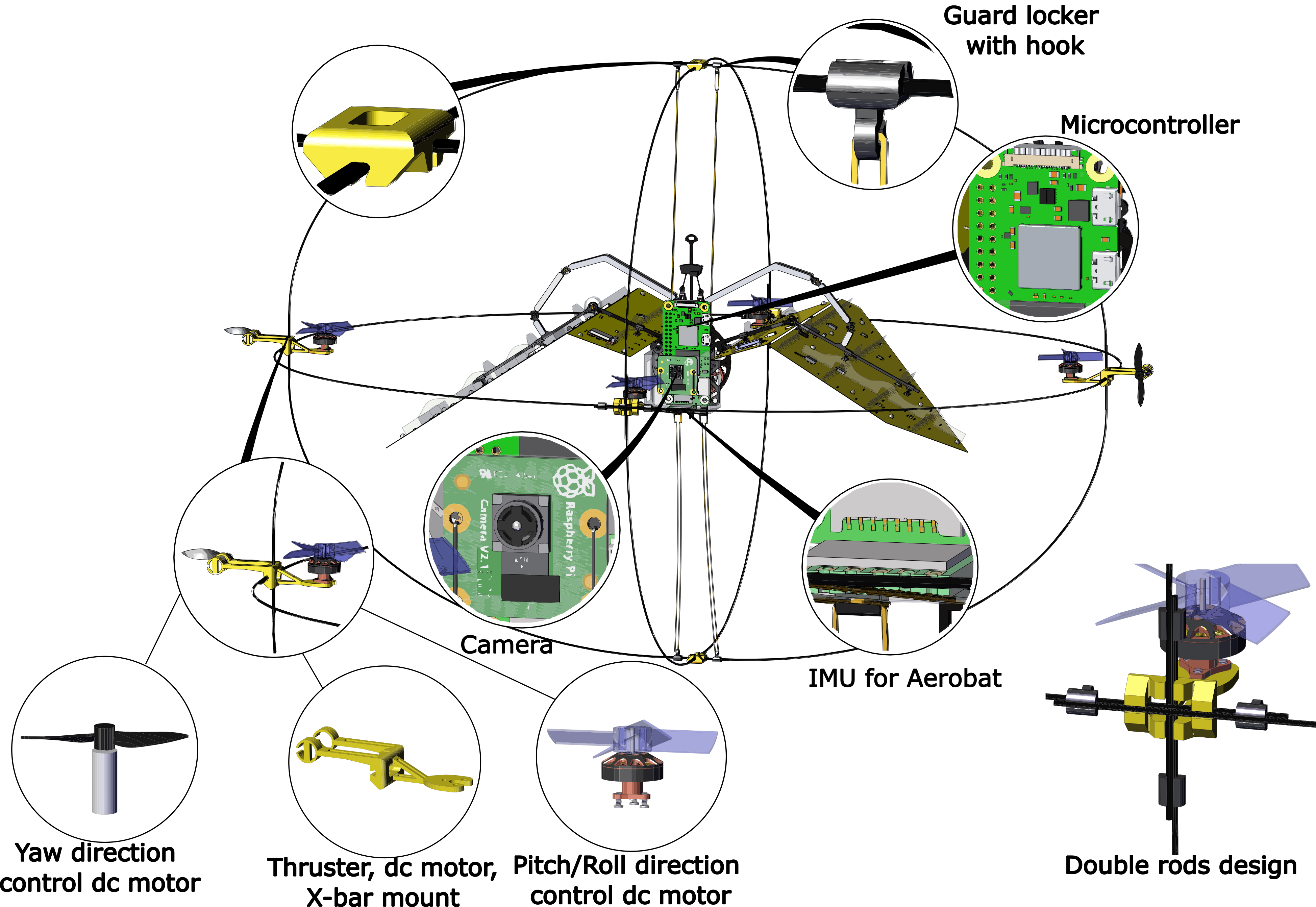}
    \caption{Illustrates the Guard design}
    \label{fig:Guard_detail_view}
\end{figure}

\begin{figure}[htp]
    \centering
    \includegraphics[width=1\linewidth]{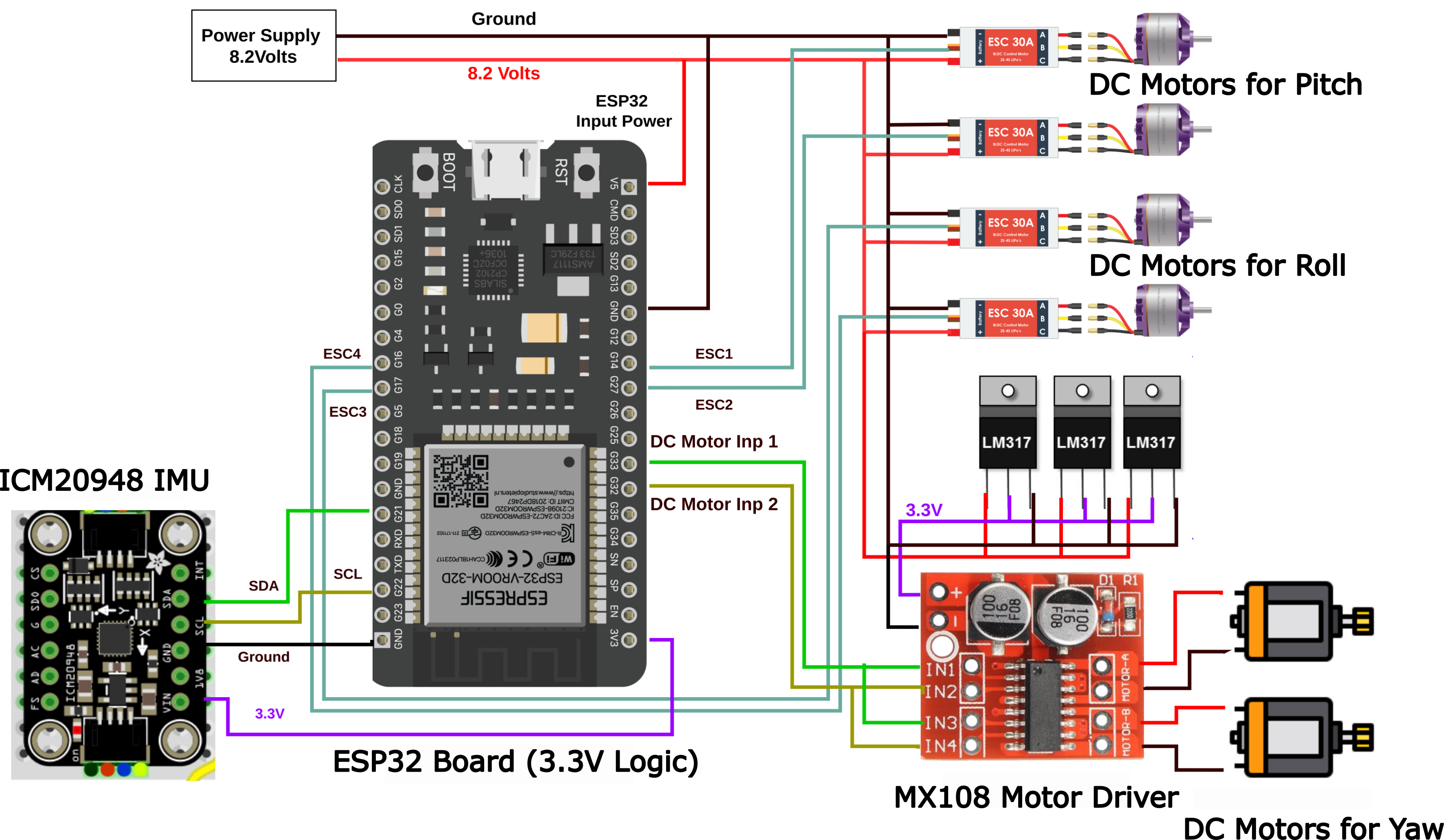}
    \caption{Electronic components on the guard}
    \label{fig:Guard_ele_detail_view}
\end{figure}

In the development of the Guard, material properties play a crucial role in ensuring optimal protection for the Aerobat platform. We chose carbon fiber rods for their low weight, high strength and durability, and ability to deform without breaking, providing a damping effect. 3D-printed PLA was chosen for mounts and connectors for the inexpensive flexibility of design they provide.

Considering that the guard needs to possess impact resistance while minimizing weight, 8 carbon fiber rods with a diameter of 1mm and a length of 400mm and 4 carbon fiber rods with a diameter of 1mm and a length of 300mm were chosen. At the connection points of the carbon fiber rods, custom-designed PLA-printed connectors were used to form three elliptical loops as shown in Fig. \ref{fig:Guard_detail_view}. The rod connection design incorporates a double-rod structure with a 3D-printed X-bar mount. The increased force-bearing surface reduces deformation caused by the Aerobat's weight and enhances the tightness of the x-rod. This results in further improvement of the Guard's structural strength and stability. To ensure secure connections, these 3D-printed connectors are affixed using glue, preventing any shifts or changes in the guard's size following collisions.

\section{Suspension System}

\begin{figure}[htp]
    \centering
    \includegraphics[width=1.0\linewidth]{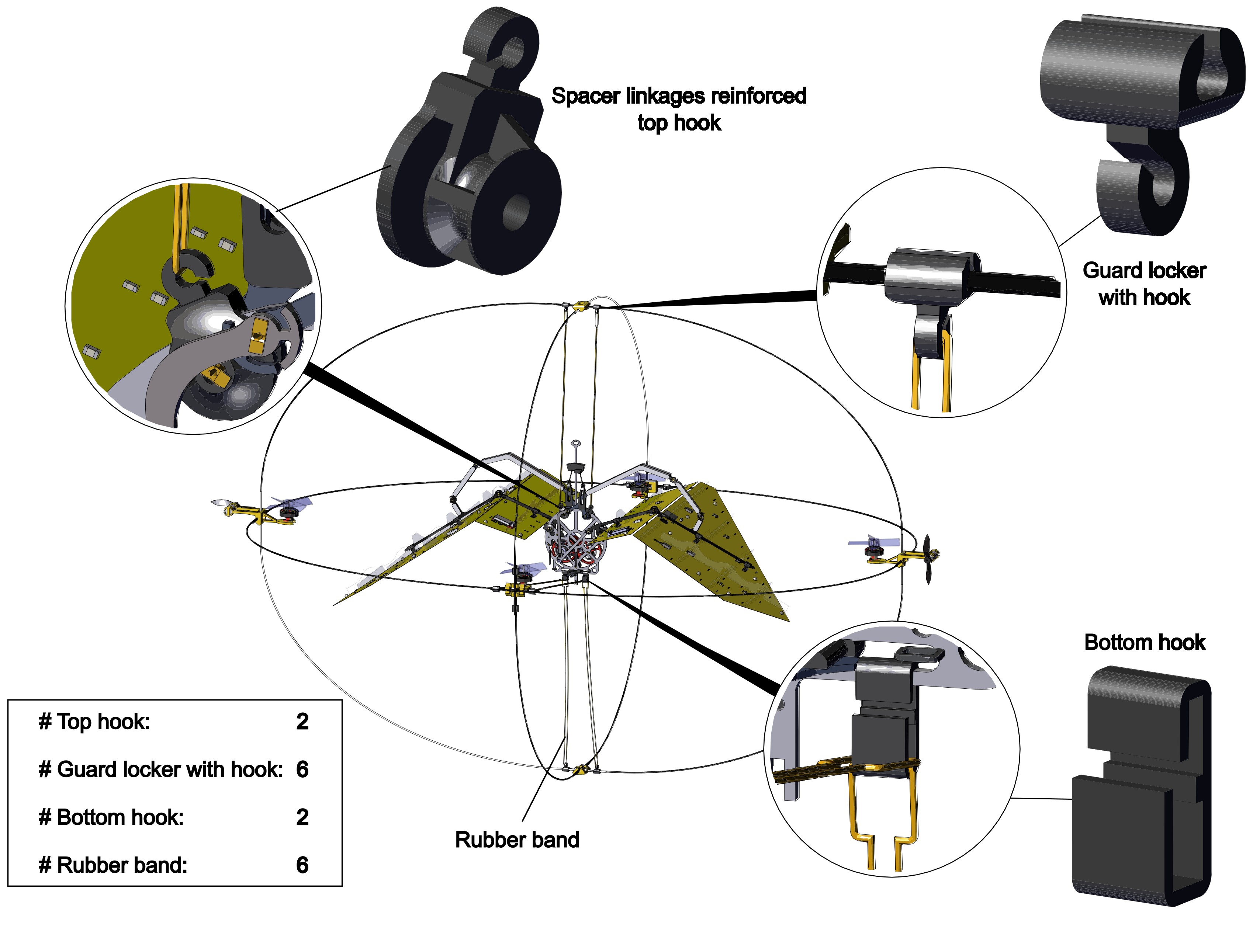}
    \caption{The CAD model of Aerobat installation inside the Guard}
    \label{fig:Aerobat_Guard_view}
\end{figure}

The attachment of Aerobat to its guard can be characterized by either rigid or compliant connections. To mitigate disturbances arising from guard movements, the Aerobat employs a flapping robot isolated with multiple elastic elements, as depicted in Fig.~\ref{fig:Aerobat_Guard_view}. For the integration of the Guard with the Aerobat platform, we utilize rubber bands. Specifically, rubber bands with an elastic coefficient 
of k=45.0N/m secure the main body of the Aerobat to the guard. The connections between the guard and the Aerobat via these rubber bands are accomplished using PLA-printed hooks.

\section{Active Stabilization}

With the Guard, our objective extends beyond providing passive protection for the Aerobat. We also strive to explore its potential to enhance flight control effectiveness. In this study, our goal is to achieve stable hovering maneuvers with the Aerobat. As the actuation framework for the Aerobat is currently under design, we are exploring alternative closed-loop operation options \cite{ramezani2022aerobat}by incorporating external position and orientation compensators through the use of a guard.

To address the challenges associated with hovering and lay the groundwork for closed-loop control implementation, we integrated an active stabilization feature into the guard design, as depicted in Fig.~\ref{fig:Guard_detail_view}. This guard incorporates six compact DC motors (comprising four brushless and two coreless types) to ensure stabilization of the roll, pitch, yaw, and x-y-z positioning of the robot. While these thrusters can counterbalance their weight and offset the additional mass of the guard and its electronics, they lack the force to lift the system autonomously. Thus, the Aerobat's lift force remains indispensable for hovering. For yaw dynamics stabilization, the guard features two lightweight motors strategically positioned at the termini of its elongated axis, as illustrated in Fig \ref{fig:Aerobat_Guard_view}. Given their extended arm placement, these motors can maintain the roll, pitch, and yaw angles without being overly robust, ensuring minimal weight addition.


The active stabilization provided by the Guard enables Aerobat to evaluate its thrust generation capabilities and gain partial control over the combined system's position and heading in incremental steps. As Aerobat's controls undergo refinement and more accurate models are developed, the Guard's involvement in active stabilization will be gradually reduced, allowing Aerobat to assume full control of its untethered flights and flight maneuvers. This gradual transition will enable the assessment of Aerobat's reliability and ensure its capacity to independently manage its flight operations.

Equipped with motion capture markers, the Guard facilitates closed-loop control and wirelessly communicates with an Optitrack motion capture system. The Guard features an onboard ESP32 microcontroller, as shown in Fig. \ref{fig:Guard_detail_view}, which serves as its primary flight controller. The ESP32 boasts a dual-core Tensilica LX6 microprocessor with clock speeds of up to 240 MHz, along with integrated WiFi connectivity for transmitting data from OptiTrack. It also controls the six electronic speed controllers. To estimate the Guard's orientation, an IMU (ICM-20948) is used, complementing the OptiTrack position and orientation data.

\chapter{Modeling and Control}
\section{Modeling}
\label{sec:modeling}

We conceptualize the dynamics between the guard and the Aerobat as interactions within a multi-agent framework. It is assumed that the guard remains unaware of the Aerobat's states.

\begin{figure*}
\vspace{0.08in}
    \centering
    \includegraphics[width=1\linewidth]{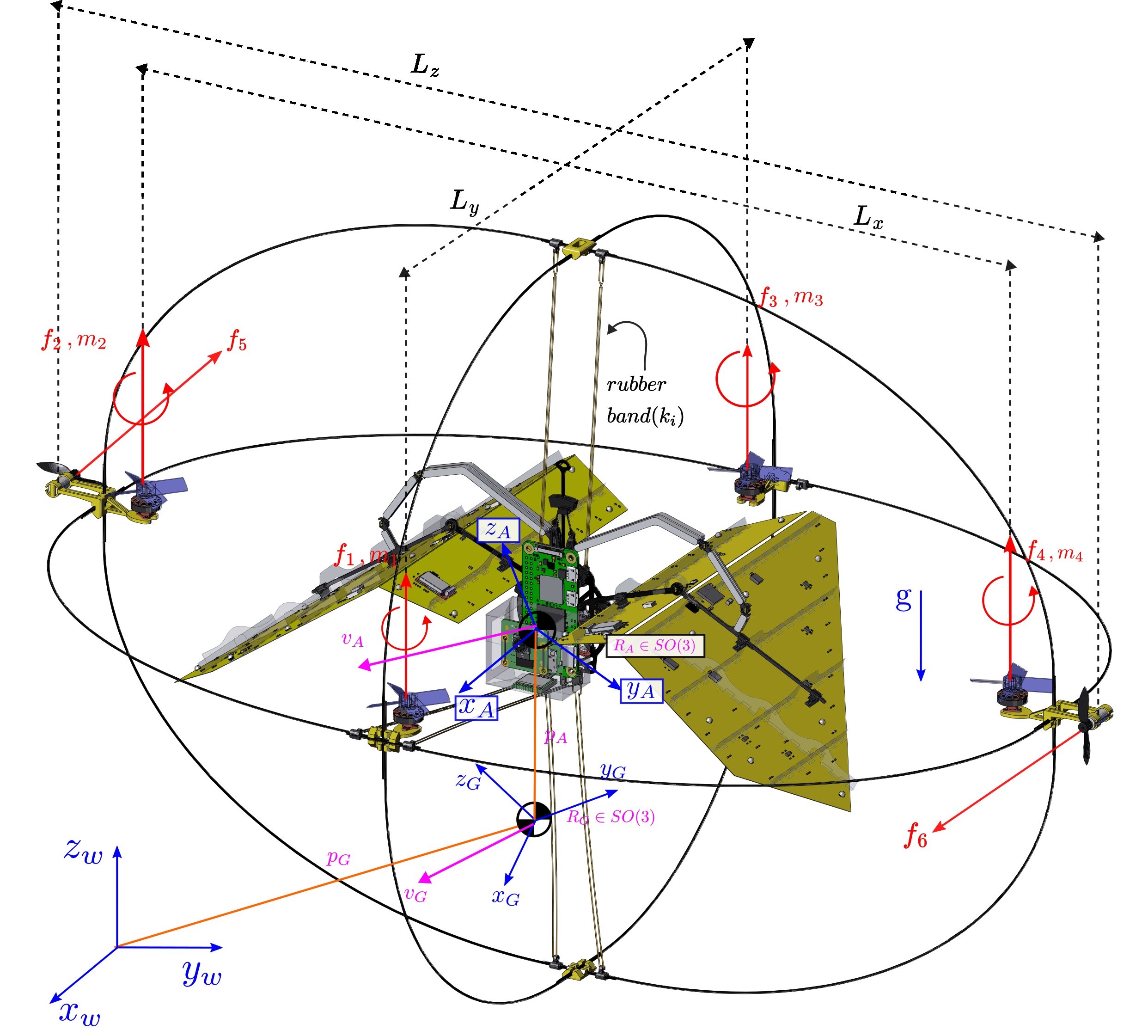}
    \caption{Illustrates Aerobat-guard free-body-diagram.}
    \label{fig:guard-fbd}
\vspace{-0.08in}
\end{figure*} 

\begin{figure}
\vspace{0.08in}
    \centering
    \includegraphics[width=0.5\linewidth]{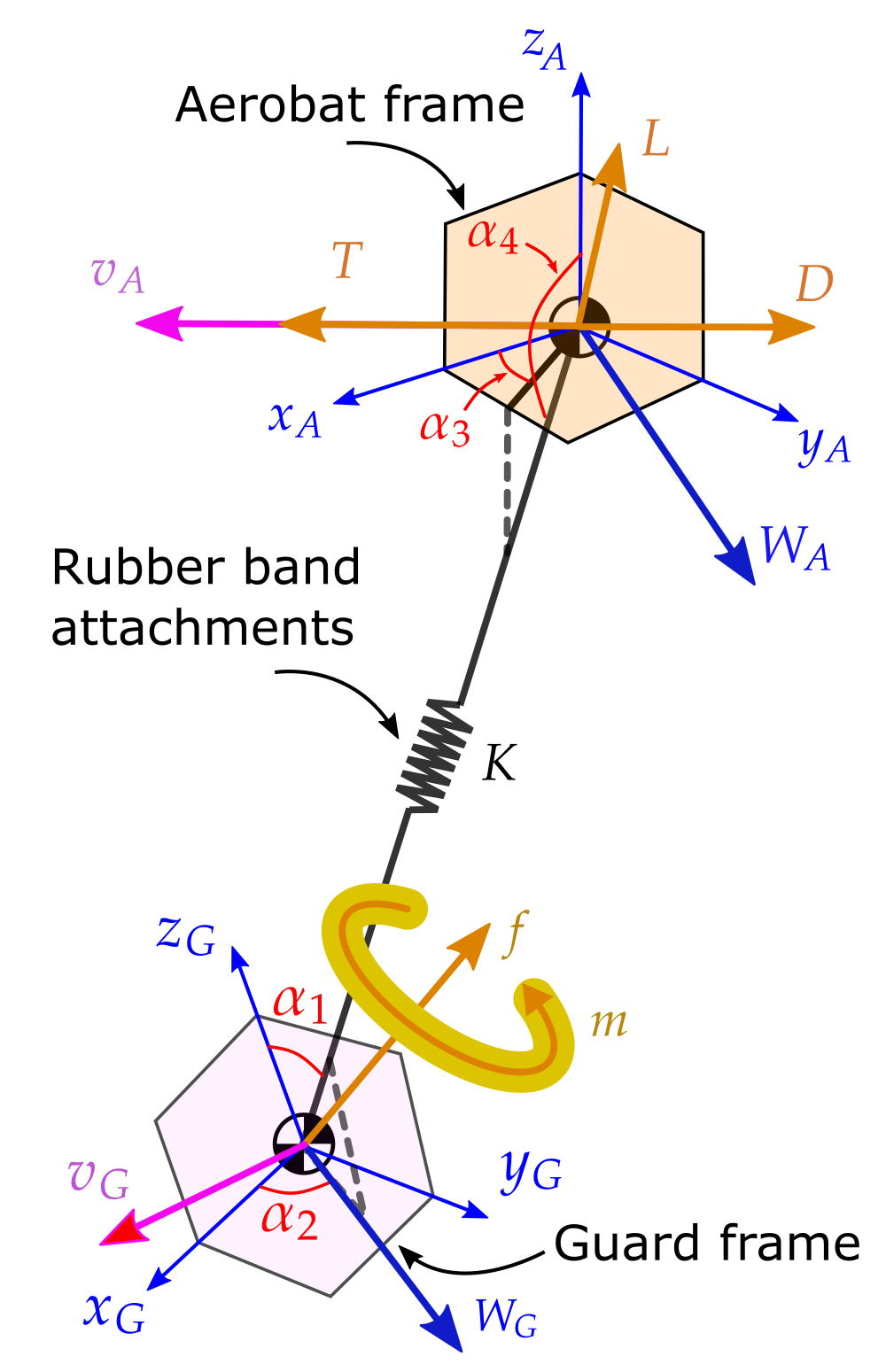}
    \caption{Shows the reduced-order-model (ROM) considered to describe Aerobat's bounding flight. $T$, $L$, and $D$ denote the thrust, lift, and drag forces that are not observable to the guard. Instead, they are added as an extended state ($x_3$ in Eq.~\ref{eq:observer}) to the guard dynamics and its estimated value used to control guard-Aerobat platform.}
    \label{fig:roms}
\vspace{-0.08in}
\end{figure}

For control design, we employ a reduced-order model (ROM) of the guard-Aerobat platform, as depicted in Fig.~\ref{fig:roms}. This ROM encompasses 14 degrees of freedom (DoFs): 6 DoFs representing the position and orientations of the guard (which are fully observable) and an additional 8 DoFs for the Aerobat. The latter includes 3 DoFs for position, 3 DoFs for orientations relative to the guard, and 2 DoFs for wing joint angles, accounting for the symmetry in the wings. This model is articulated using $q$, the vector representing configuration variables.

Consider the guard's position $p_G$, orientation (Euler angles) $q_G=[q_x,q_y,q_z]^\top$, body-frame angular velocity vector $\omega_G$, mass $m_G$, and inertia $J_G$. The body-frame forces $f_i$ and moments $m_i$ acting on the guard are:
\begin{equation}
\begin{aligned}
& f=f_1+f_2+f_3+f_4+f_5+f_6+f_e \\
& m_x=L_x\left(f_4-f_2\right)+m_{e,x} \\
& m_y=L_y\left(f_3-f_1\right)+m_{e,y} \\
& m_z=L_z\left(f_6-f_5\right)+m_{e,z}
\end{aligned}
    \label{eq:body-force-moment}
\end{equation}
\noindent where $f_i,~~i=1-6$ are roll, pitch, and yaw compensators as shown in Fig.~\ref{fig:guard-fbd}. $L_i$ is shown in Fig.~\ref{fig:guard-fbd}. In Eq.~\ref{eq:body-force-moment}, $f_e$ denotes the rubber bands' pretension forces. In Eq.~\ref{eq:body-force-moment}, we postulate that the lift, drag, and thrust forces generated by the Aerobat are transmitted to the guard via the tension forces in the rubber bands. While this supposition might be oversimplified, given the intricate wake interactions between the Aerobat's wings and the guard's propellers, it facilitates the independent modeling of these two systems using established methodologies.
The body forces are mapped to the inertial-frame force by
\begin{equation}
\left[\begin{array}{c}
F_x \\
F_y \\
F_z
\end{array}\right]=R_G^0(q_G)\left[\begin{array}{c}
0 \\
0 \\
f
\end{array}\right]
\end{equation}
\noindent This force is described in the world frame using $R_G^0$. The general equations of motion of the guard while interacting with Aerobat are given:
\begin{equation}
\Sigma_{Guard}:\left\{\begin{array}{l}
\ddot p_G= -g[0,0,1]^\top+\frac{1}{m_G}F\\
\dot R_G^0= R_G^0 \hat\omega_G \\
J_G \dot\omega_G+\omega_G \times J_G \omega_G =m 
\end{array}\right.
    \label{eq:guard-modell}
\end{equation}
\noindent where $g=9.8~ms^{-2}$. 
The equations of motion of Aerobat are given by
\begin{equation}
\Sigma_{Aerobat}:\left\{\begin{array}{l}
[\ddot p_A^\top,\ddot q_A^\top]^\top = -D_u^{-1}\Big( D_{ua} \dot a(t)-H_u+J^\top y\Big)\vspace{0.1in}\\
\dot \xi=\Pi_1(\xi)\xi+\Pi_2(\xi)a(t) \vspace{0.1in}\\
y=\Pi_3(\xi)\xi+\Pi_4(\xi)a(t)
\end{array}\right.
    \label{eq:aerobat-full-dynamics}
\end{equation}
\noindent where $\Pi_i$ and $\xi$ are the aerodynamic model parameters and fluid state vector. $y$ is the output of the aerodynamic model, that is, the aerodynamic force. $J$ is the Jacobian matrix. $a(t)=[\dots a_i(t)\dots]^\top$ denotes wings' joint trajectories. $p_A$ and $q_A$ denote the position and orientation of Aerobat with respect to the guard and are given by
\begin{equation}
    \begin{aligned}
        p_A = R_A^G(q_A)[0,0,1]^\top\\
        q_A = [\alpha_3,\alpha_4]^\top
    \end{aligned}
\end{equation}
\noindent Please see Fig.~\ref{fig:guard-fbd} for more information about $\alpha_i$. In Eq.~\ref{eq:aerobat-full-dynamics}, $D_u$, $D_{ua}$, and $H_u$ are the block matrices from partitioning Aerobat's full-dynamics. This partitioning dissects the full-dynamics inertia matrix $D(q)$, Coriolis matrix $C(q,\dot q)$, and conservative potential forces (gravity and rubber bands) $G(q)$ corresponding to the underactuated (position and orientation) $q_u=[p_A^\top,q_A^\top]^\top$ and actuated (the wings joints) $a(t)$ are separated. 

The force and moments $f_e$ and $m_e$ in Eq.~\ref{eq:body-force-moment} are obtained as elastic conservative forces. To do this, the total potential energy $V(q)$ is used to obtain $H=C(q,\dot q)\dot q + G(q)$ in Aerobat's model.  $V(q)$ is given given by
\begin{equation}
    V(q)=\frac{1}{2}K(p_G-p_A)^\top(p_G-p_A) \\+ m_Ag\Big(p_{G,z}+R_A^0p_{A,z}\Big)
    \label{eq:total-potential-energy}
\end{equation}
\noindent where $K$ denotes the rubber bands' elastic coefficient and $m_A$ is Aerobat's total mass. In Eq.~\ref{eq:aerobat-full-dynamics}, the interactions between the Aerobat and its fluidic environment are represented using our experimentally validated unsteady-quasi-static models, assessed at discrete points along the wing span. The aerodynamic model outputs $y=[y1,\dots,y_m]^\top$, where $y_i$ is the external aerodynamic force at i-th strip is governed by the state-space model $\dot \xi$. We briefly cover this model; however, the reader is referred to \cite{sihite2022unsteady} for more details on the model derivations. 

\subsection{Brief Overview of Unsteady Aerodynamic State-Space Model $\dot \xi=A(\xi)\xi+B(\xi)a(t)$}

We superimpose horseshoe vortices on and behind the wing blade elements to calculate lift and drag forces. Consider the time-varying circulation value at $s_i$, the location of the i-th element on the wing, denoted by $\Gamma_i(t)$. The circulation can be parameterized by truncated Fourier series of $n$ coefficients. $\Gamma_i$ is given by 
\begin{equation}
\begin{aligned}
    \Gamma_i(t) = a^\top(t)
    \begin{bmatrix}
    \sin(\theta_i)\\
    \vdots\\
    \sin(n\theta_i)\\
    \end{bmatrix},
\end{aligned}
    \label{eq:circulation}
\end{equation}
\noindent where $a=[a_1,\dots,a_n]^\top$ are the Fourier coefficients and $\theta_i=\arccos(\frac{s_i}{l})$ ($l$ is the wingspan size). From Prandtl's lifting line theory, additional circulation-induced kinematics denoted by $y_{\Gamma}$ are considered on all of the points $p_i$. These circulation-induced kinematics are given by
\begin{equation}
\begin{aligned}
    y_{\Gamma} = 
    \begin{bmatrix}
    1&\frac{\sin{2\theta_1}}{\sin{\theta_1}}&\dots&\frac{\sin{n\theta_1}}{\sin{\theta_1}}\\
    1&\frac{\sin{2\theta_2}}{\sin{\theta_2}}&\dots&\frac{\sin{n\theta_2}}{\sin{\theta_2}}\\
    \vdots\\
    1&\frac{\sin{2\theta_n}}{\sin{\theta_n}}&\dots&\frac{\sin{n\theta_n}}{\sin{\theta_n}}\\
    \end{bmatrix}a(t)
    \label{eq:induced-kin}
\end{aligned}
\end{equation}
Now, we utilize a Wagner function $\Phi(\tau) = \Sigma_{k=1}^2\psi_k \exp{(-\frac{\epsilon_k}{c_i} \tau)}$, where $\psi_k$, $\epsilon_k$ are some scalar coefficients and $\tau$ is a scaled time to compute the aerodynamic force coefficient response $\beta_{i}$ associated with the i-th blade element. The kinematics of i-th element using Eqs.~\ref{eq:aerobat-full-dynamics} and \ref{eq:induced-kin} are given by 
\begin{equation}
y'_{1,i} = y_{1,i} + y_{\Gamma,i}.    
\end{equation}
\noindent Following Duhamel's integral rule, the response is obtained by the convolution integral given by
\begin{equation}
    \beta_i = y'_{1,i}\Phi_0 + \int_0^t\frac{\partial \Phi(t-\tau)}{\partial \tau}y'_{1,i}d\tau \\
\label{eq:lift_coeff_wagner}
\end{equation}
\noindent where $\Phi_0=\Phi(0)$. We perform integration by part to eliminate $\frac{\partial \Phi}{\partial \tau}$, substitute the Wagner function given above in \eqref{eq:lift_coeff_wagner}, and employ the change of variable given by $z_{k,i} (t) = \int_{0}^{t} \exp{(-\frac{\epsilon_k}{c_i}(t-\tau))} y'_{1,i} d\tau$, where $k\in\{1,2\}$, to obtain a new expression for $\beta_i$ based on $z_{k,i}$  
\begin{equation}
\beta_i =
y'_{1,i}\Phi_0+
\begin{bmatrix}
\psi_1\frac{\epsilon_1}{c_i}&\psi_2\frac{\epsilon_2}{c_i}
\end{bmatrix}
\begin{bmatrix}
z_{1,i}\\
z_{2,i}
\end{bmatrix}
\label{eq:?1}
\end{equation}
The variables $a$, $z_{1,i}$, and $z_{2,i}$ are used towards obtaining a state-space realization that can be marched forward in time. Using the Leibniz integral rule for differentiation under the integral sign, unsteady Kutta–Joukowski results $\beta_i=\frac{\Gamma_i}{c_i}+\frac{d\Gamma_i}{dt}$, and \eqref{eq:circulation}, the model that describes the time evolution of the aerodynamic states is obtained
\begin{equation}
\Sigma_{Aero,i}:\left\{
\begin{aligned}
    A_i\dot{a} &= -B_ia + C_iZ_i  + \Phi_0y'_{1,i}\\
    \dot{Z}_i &= D_iZ_i + E_iy'_{1,i}\\
\end{aligned}
\right.
    \label{eq:aerodynamic-ss-model}
\end{equation}
\noindent where $Z_i$, $A_i$, $B_i$, $C_i$, $D_i$, and $E_i$ are the state variable and matrices corresponding to the i-th blade element. They are given by
\begin{equation}
\begin{aligned}
    Z_i &= 
    \begin{bmatrix}
    z_{1,i} & z_{2,i}
    \end{bmatrix}^\top,\\
    A_i &= 
    \begin{bmatrix}
    \sin{\theta_i} & \sin{2\theta_i} & \dots & \sin{n\theta_i} 
    \end{bmatrix},\\
    B_i &= A_i/c_i,\\
    C_i &= 
    \begin{bmatrix}
    \frac{\psi_1\epsilon_1}{c_i} & \frac{\psi_2\epsilon_2}{c_i}
    \end{bmatrix},\\
    D_i &= 
    \begin{bmatrix}
    \frac{-2\epsilon_1}{c_i} & 0\\
    0 & \frac{-2\epsilon_2}{c_i}
    \end{bmatrix},\\
    E_i &=
    \begin{bmatrix}
    2-\exp{\frac{\epsilon_1 t}{c_i}} & 2-\exp{\frac{\epsilon_2 t}{c_i}}
    \end{bmatrix}^\top.
\end{aligned}
    \label{eq:?2}
\end{equation}
\noindent We define the unified aerodynamic state vector, used to describe the state space of \eqref{eq:aerobat-full-dynamics}, as following $\xi = [a^\top, Z^\top]^\top$, where $Z=[Z^\top_1,\dots,Z^\top_n]^\top$.

\section{Control}

The full dynamics of the guard with Aerobat sitting inside are given by
\begin{equation}
\Sigma_{FullDyn}\left\{\begin{array}{l}
\dot{x}_1=x_2 \\
\dot{x}_2=g_1+g_2u+g_3x_3 \\
\dot{x}_3=G(t) \\
z=x_1
\end{array}\right.
    \label{eq:extended-state-model}
\end{equation}
\noindent where $x_1=[p_G^\top,q_G^\top]^\top$, the nonlinear terms $g_i$ are given by Eqs.~\ref{eq:guard-modell} and \ref{eq:aerobat-full-dynamics}, $u=[\dots f_i \dots]$ from Eq.~\ref{eq:body-force-moment}, and $x_3=y$ from Eq.~\ref{eq:aerobat-full-dynamics}. As it can be seen, the model is extended with another state $x_3$ because $y$ (aerodynamic force) can be written in the state-space form given by Eq.~\ref{eq:aerobat-full-dynamics}. 

We now turn our attention to the control presented in Eq.~\ref{eq:extended-state-model}, operating under the assumption that $u$ is derived solely from observations of $z=x_1$. Though the time-varying term $G(t)$ - which represents the dynamics of $y$ - is inherently nonlinear, an efficient model for $G(t)$ \cite{sihite2022unsteady} is available. In \cite{sihite2022unsteady}, we introduced an aerodynamic model that accurately predicts the external aerodynamic forces acting on Aerobat. 

Using this model, we have developed a state observer for $x_3$. This approach enhances the feedback $u=Kx_2$, where $K$ denotes the control gain so that $\dot x_2$ remains bounded and stable. For a clearer definition of the estimated states $\hat x_i$, please refer to \cite{sihite2022unsteady}:
\begin{equation}
\begin{aligned}
& \hat{x}_1=\hat{x}_2-\beta_1\left(\hat{x}_1-x_1\right) \\
& \hat{x}_2=g_1+g_2 u+g_3\hat{x}_3-\beta_2\left(\hat{x}_1-x_1\right) \\
& \hat{x}_3=-\beta_3\left(\hat{x}_1-x_1\right)
\end{aligned}
    \label{eq:estimated-states}
\end{equation}
\noindent where $\beta_i$ is the observer gains. Now, we define the error $e_i=\hat x_i- x_1$ for $i=1,2,3$. The following observer model is found
\begin{equation}
\left[\begin{array}{c}
\dot e_1 \\
\dot e_2\\
\dot e_3
\end{array}\right]=\left[\begin{array}{ccc}
-\beta_1 & I & 0 \\
-\beta_2 & 0 & g_3 \\
-\beta_3 & 0 & 0
\end{array}\right]\left[\begin{array}{c}
e_1 \\
e_2 \\
e_3
\end{array}\right]+\left[\begin{array}{c}
0 \\
0 \\
-I
\end{array}\right] G(t)
    \label{eq:observer}
\end{equation}
\noindent where ... The gains $\beta_i$ for the observer given by Eq.~\ref{eq:observer} can be obtained if upper bounds for $G$ and $g_2$ can be assumed. We have extensively studied $g_1$, $g_2$ and $g_3$ terms in Aerobat's model in past and ongoing efforts. Based on the bounds for $\|g_1\|$, $\|g_2\|$, and $\|g_3\|$, we tuned the observer. The controller used for the bounding flight is given by
\begin{equation}
u = g_2^{-1}\Big(u_0 -g_1 - g_3\hat x_3\Big) 
    \label{eq:controller}
\end{equation}
\noindent where $u_0=Kx_2$.

\chapter{Experiment}


\renewcommand\floatpagefraction{.9}
\renewcommand\topfraction{.9}
\renewcommand\bottomfraction{.9}
\renewcommand\textfraction{.1}
\setcounter{totalnumber}{50}
\setcounter{topnumber}{50}
\setcounter{bottomnumber}{50}



\section{Aerobat flapping with guard stabilization}
\begin{figure*}
\vspace{0.08in}
    \centering
    \includegraphics[width=1\linewidth]{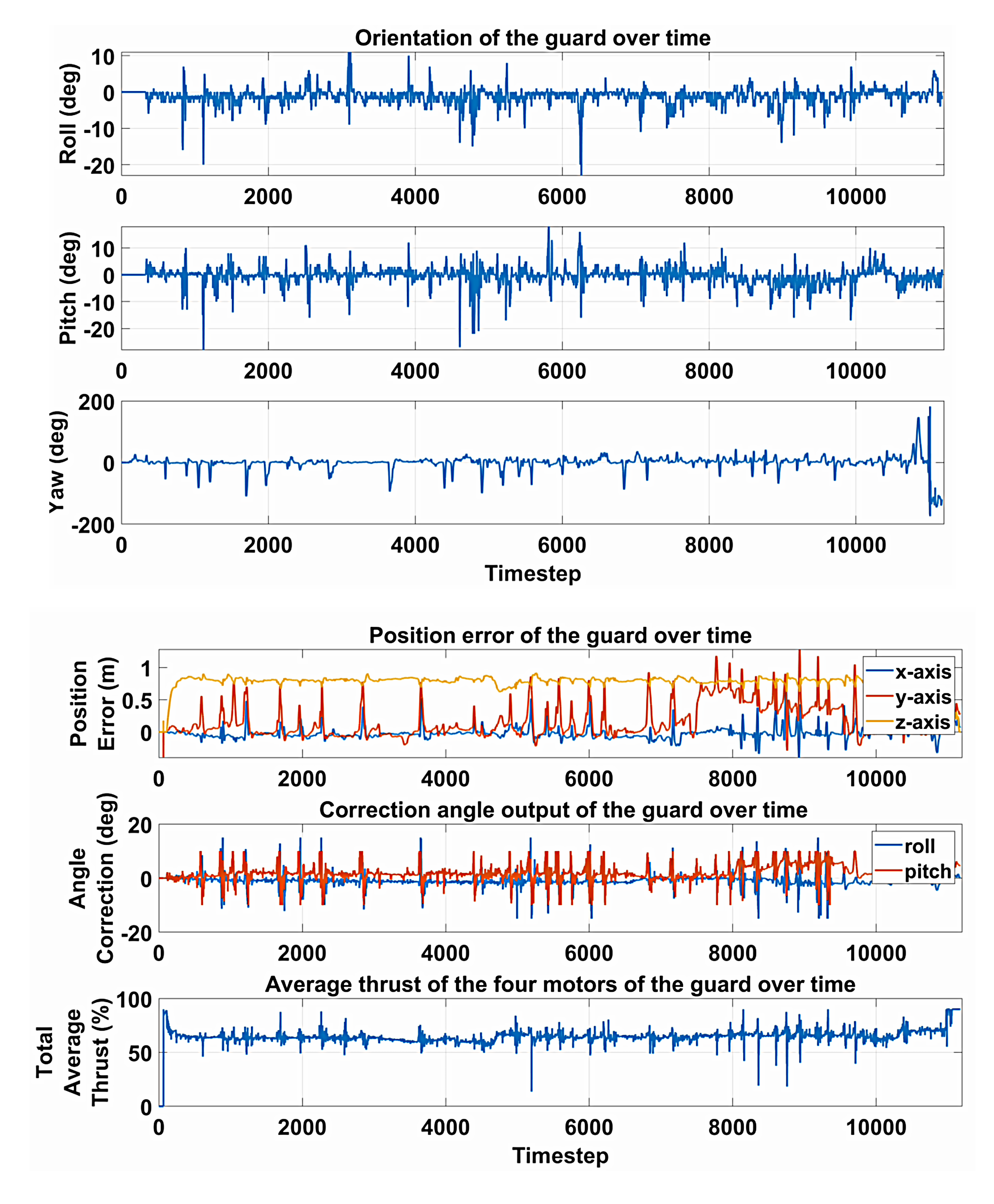}
    \caption{Shows the tracking performance of the controller based on estimated generalized inertial $g_1^{-1}f$ and aerodynamics $g_1^{-1}g_2\hat x_3$ to maintain a fixed position and orientation of the guard.}
    \label{fig:optitrack}
\vspace{-0.08in}
\end{figure*}

\begin{figure}
\vspace{0.08in}
    \centering
    \includegraphics[width=1\linewidth]{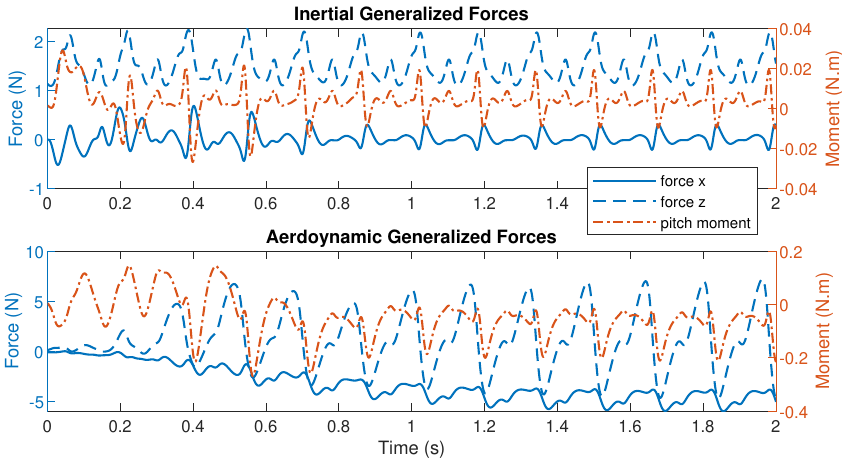}
    \caption{Shows estimated generalized inertial $g_2^{-1}g_1$ and aerodynamics $g_2^{-1}g_3\hat x_3$ contributions based on knowledge on the boundedness of \textbf{$\|g_2\|$} and \textbf{$\|G(t)\|$}.}
    \label{fig:gen_force}
\vspace{-0.08in}
\end{figure}

The determination of the control command $u=[f_1,\dots,f_6]^\top$ was accomplished by using inferred values of the augmented state variable $x_3$ during stationary flight within the RISE arena in a controlled manner. The state variable $x_2$, encapsulating the positional coordinates, orientations, and velocities of the custodial entity, was obtained with the help of our OptiTrack motion capture and onboard IMU. The dynamics of unsteady aerodynamics, as proposed in \cite{sihite2022unsteady}, were employed to identify the constraints on the parameter $G(t)$ during the stationary flight phase. Established on insights from the unsteady aerodynamic model, $\|G(t)\|$ was chosen for the stationary flight regime. The associated estimations $g_1^{-1}g_2\hat x_3$, representing contributions from generalized aerodynamic forces during stationary flight, are depicted in Fig.~\ref{fig:gen_force}. Likewise, the derived estimations pertaining to $g_1^{-1}f$, indicating contributions from generalized inertial dynamics and are presented in Fig.~\ref{fig:gen_force}, and were also used to derive control command vector u. The efficacy of the control strategy in achieving stabilization across roll, pitch, yaw, and x-y-z positional coordinates is also depicted in Figs ~\ref{fig:optitrack}.


\section{Wingtip trajectory experiment}To assess the efficacy of wing flapping, the AeroBat's left wingtip position was monitored using the OptiTrack System. For this study, three markers were placed on the left wingtip, which was then defined as a body within the OptiTrack software using its body tracking feature. Following the flapping test, the software recorded the wingtip's position and rotation data. This information was later visualized in MATLAB, showcasing the wingtip's trajectory over time.

\begin{figure}
\vspace{0.08in}
    \centering
    \includegraphics[width=1\linewidth]{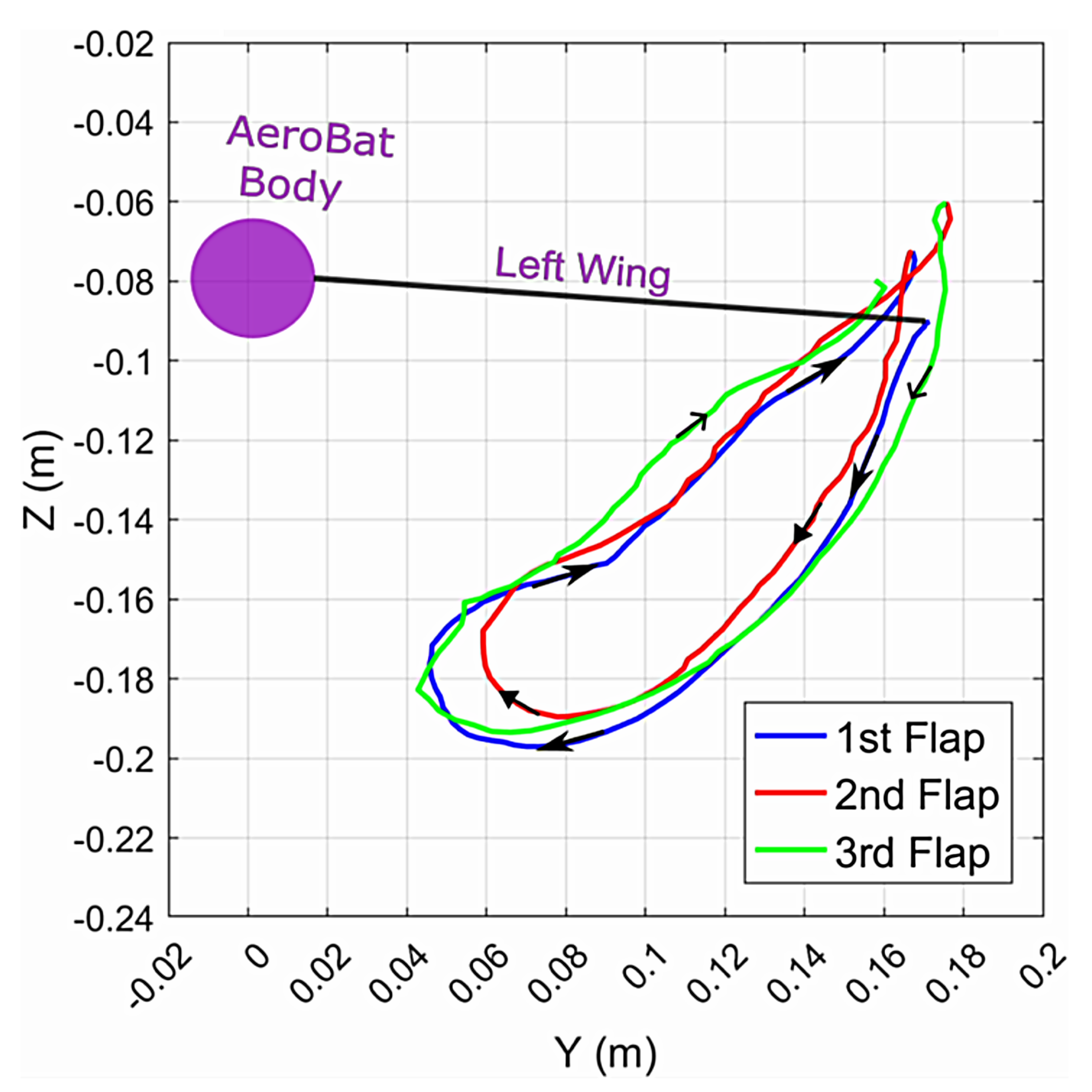}
    \caption{Aerobat wingtip trajectory}
    \label{fig:wingtip track}
\vspace{-0.08in}
\end{figure}

Fig.~\ref{fig:wingtip track}illustrates the wingtip's trajectory during three flapping motions, depicted in the Y vs Z coordinates relative to the AeroBat's body. The prominent pink circle and the black line represent the AeroBat's body and left wing, respectively. The three trajectories closely resemble each other, attesting to the effective installation of the PCB wing. However, minor variances in these paths could be attributed to the flexible PCB wing, which may experience slight elastic deformations when flapping at high velocities.

\chapter{Conclusion and Future Work}
\label{chap:conclude}


\section{Conclusions}
\label{sec:conclusion}

The primary motivation of this MS thesis revolves around integrating polyimide flexible PCB into the Northeastern Aerobat, emphasizing the potential and benefits of combining traditional engineering approaches with bio-inspired designs. This integration offers a lightweight solution and harnesses the advantages of flexibility, enhancing the Aerobat's capability for safe hovering in confined spaces.

Bio-inspired drones are steadily establishing themselves as pivotal tools in a variety of sectors, from wildlife monitoring to ensuring residential safety. Mirroring the sophisticated noise-reduction techniques inherent in birds and insects, such drones can operate without causing disturbances, a critical factor for tasks like surveillance and wildlife observation. However, the challenges traditional drones face, especially in confined spaces, highlight the importance of innovative solutions. 

In the design evolution of the Aerobat, a prominent challenge emerged: the risk of the elbow joint damaging the flexible PCB during wing retraction by intruding upon the wing membrane. To circumvent this, a meticulously engineered safe zone was established between the wing membrane and arm. Yet, this solution inadvertently introduced the potential for compressive or tensile stresses, attributed to the buckling effects of laminated structures. The ingenious introduction of the living hinge structure, crafted using Onyx material, became instrumental. This structure not only compensated for the bending dynamics of the PCB but also optimized aerodynamics, epitomizing the strategic balance between protection and performance. Paired with feedback-driven stabilizers, these innovations mark significant strides in addressing inherent design challenges. 

Origami, a traditional art form, is being ingeniously woven into modern robotics, as exemplified by the electronic cage design in this research. The drastic weight reduction of approximately 76.4\% achieved with the origami-inspired cage is a testament to the fusion of old-world craftsmanship and contemporary engineering. Such designs, when coupled with the robustness of materials like polypropylene, promise not only lightweight structures but also enhanced durability.

In ensuring the Aerobat's safety, the elliptical guard design's effectiveness in dispersing impact stresses stands out. The incorporation of carbon fiber rods and custom-designed PLA-printed connectors further illustrates the delicate balance of creating a lightweight yet sturdy protective mechanism. With the guard's design leaning towards not just passive protection but active stabilization, the Aerobat edges closer to achieving independent untethered flights.

Technological integration with systems like the OptiTrack motion capture and the deployment of the ESP32 microcontroller augments Aerobat's flight control prowess. The analytic depth provided by the reduced-order model (ROM) and derived aerodynamic models substantiates the robustness of the control mechanisms incorporated.

In summary, this thesis underscores the potential of combining traditional engineering approaches with bio-inspired designs. It illustrates how innovation, inspired by nature and underpinned by cutting-edge technology like the polyimide flexible PCB, can propel aerial robots like the Aerobat to new heights. As drones become more intertwined with daily human life, advancements from studies like this will be foundational in shaping a future with safer, more adaptable, and highly efficient aerial vehicles.

\section{Future Work}
\label{sec:future work}
A primary objective in our forthcoming studies is the detailed measurement and analysis of wing membrane deformation. Given the intricacies associated with wing dynamics, we are setting our sights on leveraging the capabilities of the OptiTrack system. This system's precision and reliability will enable a comprehensive understanding of deformations, allowing for tailored improvements in wing design and functionality.

Furthermore, a critical facet of our future work involves reevaluating the role of the Guard in active stabilization. While the Guard has been instrumental in safeguarding the Aerobat during its nascent stages, the aim is to incrementally reduce its contribution to stabilization. This transition will be instrumental in enhancing the Aerobat's self-reliance, moving it closer to achieving more autonomous and untethered flights.

Simultaneously, our vision for the Aerobat is not confined to its physical design alone. We anticipate a transformative journey where the Aerobat seamlessly melds with mechanical intelligence. The foundation for this lies in the integration of electronic components on the flexible PCB. By combining the inherent agility of the Aerobat with advanced electronic intelligence, we aim to pioneer a new generation of bio-inspired drones that are not only mechanically adept but also cognitively advanced.


\printbibliography

\appendix


\printindex

\end{document}